\documentclass{article}

\usepackage[main,final,nonatbib]{neurips_2026}

\usepackage[utf8]{inputenc}
\usepackage[T1]{fontenc}
\usepackage{amsmath,amssymb,amsfonts}
\usepackage{mathtools}
\usepackage{algorithm}
\usepackage{algpseudocode}
\usepackage{graphicx}
\usepackage{subcaption}
\usepackage{booktabs}
\usepackage{hyperref}
\usepackage{cleveref}
\usepackage{xcolor}
\usepackage{enumitem}
\usepackage{amsfonts}       
\usepackage{nicefrac}       
\usepackage{microtype}
\usepackage{float}
\usepackage{microtype}      
\usepackage{xcolor}         
\usepackage{multirow}
\usepackage[numbers,sort&compress]{natbib}
\usepackage{comment}  
\usepackage{tikz}
\usepackage{amsmath,amssymb}
\usetikzlibrary{positioning, arrows.meta, shapes.geometric, fit, calc, backgrounds}

\definecolor{genBlue}{HTML}{2B7BBF}       
\definecolor{genBlueBg}{HTML}{DCEAF7}     
\definecolor{scoreRed}{HTML}{C0392B}      
\definecolor{scoreRedBg}{HTML}{F5D7D3}    
\definecolor{trainGreen}{HTML}{1E8449}    
\definecolor{trainGreenBg}{HTML}{D5F0DE}  
\definecolor{detailPurple}{HTML}{6C3483}  
\definecolor{detailPurpleBg}{HTML}{EBDEF0} 
\definecolor{secRef}{HTML}{777777}        
\definecolor{arrowGray}{HTML}{444444}     
\definecolor{loopGreen}{HTML}{27AE60}     

\definecolor{boxfill}{HTML}{ECECEC}
\definecolor{boxedge}{HTML}{B0B0B0}
\definecolor{numgrey}{HTML}{707070}
\definecolor{posgreen}{HTML}{2E8B57}
\definecolor{negred}{HTML}{C0392B}
\definecolor{vlmtext}{HTML}{3D4A7A}
\definecolor{vlmfill}{HTML}{E6ECF6}
\definecolor{vlmedge}{HTML}{B8C4DD}
\definecolor{arrowgrey}{HTML}{9E9E9E}

\newcommand{\method}{Stable-Layers}
\newcommand{\modelname}{Qwen-Image-Layered}

\newcommand{\E}{\mathbb{E}}
\newcommand{\KL}{\mathrm{KL}}
\newcommand{\vtheta}{v_\theta}

\makeatletter
\newif\ifshowchecklist
\if@preprint
  \showchecklistfalse
\else
  \showchecklisttrue
\fi
\makeatother

\usepackage{xspace}
\usepackage{booktabs}
\usepackage{caption}
\usepackage{subcaption}
\usepackage{float}
\usepackage{wrapfig}
\usepackage{multirow}
\usepackage{colortbl}
\usepackage[skins]{tcolorbox}

\definecolor{bestcol}{RGB}{254,196,79}
\newcommand{\best}[1]{\cellcolor{bestcol} \textbf{#1}}
\definecolor{secondbestcol}{RGB}{255,247,188}

\makeatletter
\DeclareRobustCommand\onedot{\futurelet\@let@token\@onedot}
\def\@onedot{\ifx\@let@token.\else.\null\fi\xspace}

\def\etal{\emph{et al}\onedot}

\makeatother

\newcommand{\inlinesection}[1]{\vspace{0.05cm} \noindent {\bf #1}}
\newcommand{\titlecaption}[2]{\caption{\textbf{#1.}\xspace#2}}

\usepackage{pifont}
\usepackage{tikz}
\usepackage{fontawesome5}
\usetikzlibrary{arrows.meta, positioning, calc}

\title{%
  \method{}: Fine-Tuning Image Layer Decomposition Models \\
  with VLM-Scored Reinforcement Learning
}
\author{
  Ciara Rowles\\
  Stability AI \and Reshinth Adithyan\\
  Stability AI \and Nikhil Pinnaparaju\\
  Stability AI \and Vikram Voleti\\
  Stability AI \and Mark Boss\\
  Stability AI
}
\date{\today}

\begin{document}
\maketitle
\begin{center}
    \centering
    \captionsetup{type=figure}
    \scalebox{0.88}{
\newlength{\tw}\setlength{\tw}{\textwidth}%
\begin{tikzpicture}[
    >=Stealth,
    every node/.style={inner sep=0pt},
    lbl/.style={inner sep=2pt, font=\small},
    bigarrow/.style={->, line width=1.8pt},
    x=\tw, y=\tw,
]


\node[anchor=center] (source) at (0.07, 0)
    {\includegraphics[width=0.13\tw]{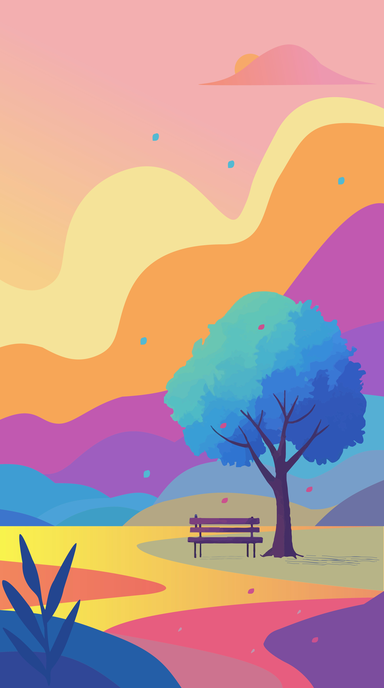}};

\coordinate (arr1-start) at ($(source.east)+(0.006,0)$);
\coordinate (arr1-end)   at (0.22, 0);
\draw[bigarrow] (arr1-start) -- (arr1-end);

\coordinate (arr1-mid) at ($(arr1-start)!0.5!(arr1-end)$);
\node[font=\small, anchor=north] at ($(arr1-mid)+(0,0.035)$)
    {\faLayerGroup};
\node[font=\scriptsize\bfseries, anchor=north, align=center] at ($(arr1-mid)+(0,-0.015)$)
    {Stable\\[-1pt]Layers};

\coordinate (grpo-land) at ($(arr1-mid)+(0, 0.04)$);

\node[anchor=center] (c1) at (0.30, 0.038)
    {\includegraphics[width=0.09\tw]{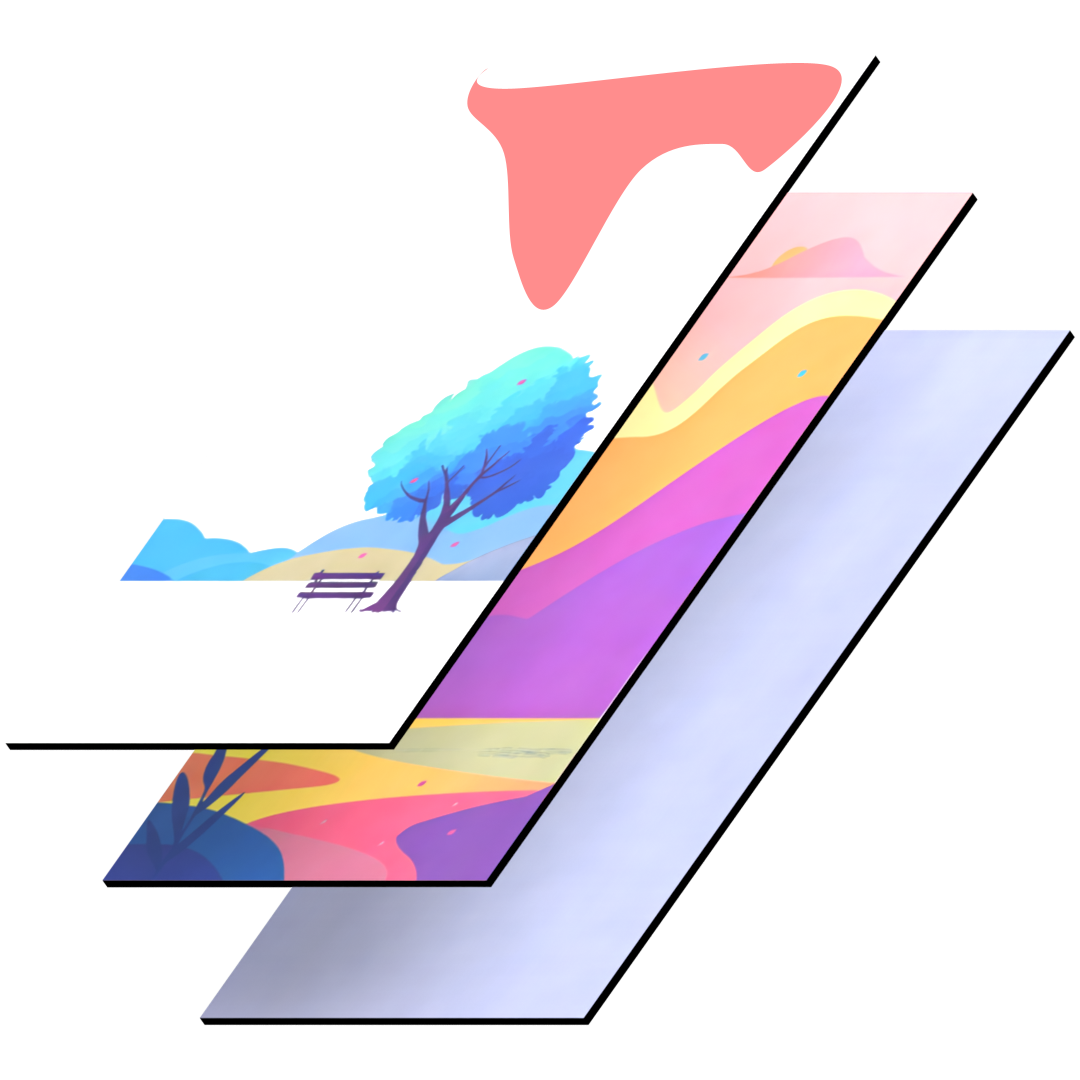}};
\node[anchor=center] (c2) at (0.40, 0.038)
    {\includegraphics[width=0.09\tw]{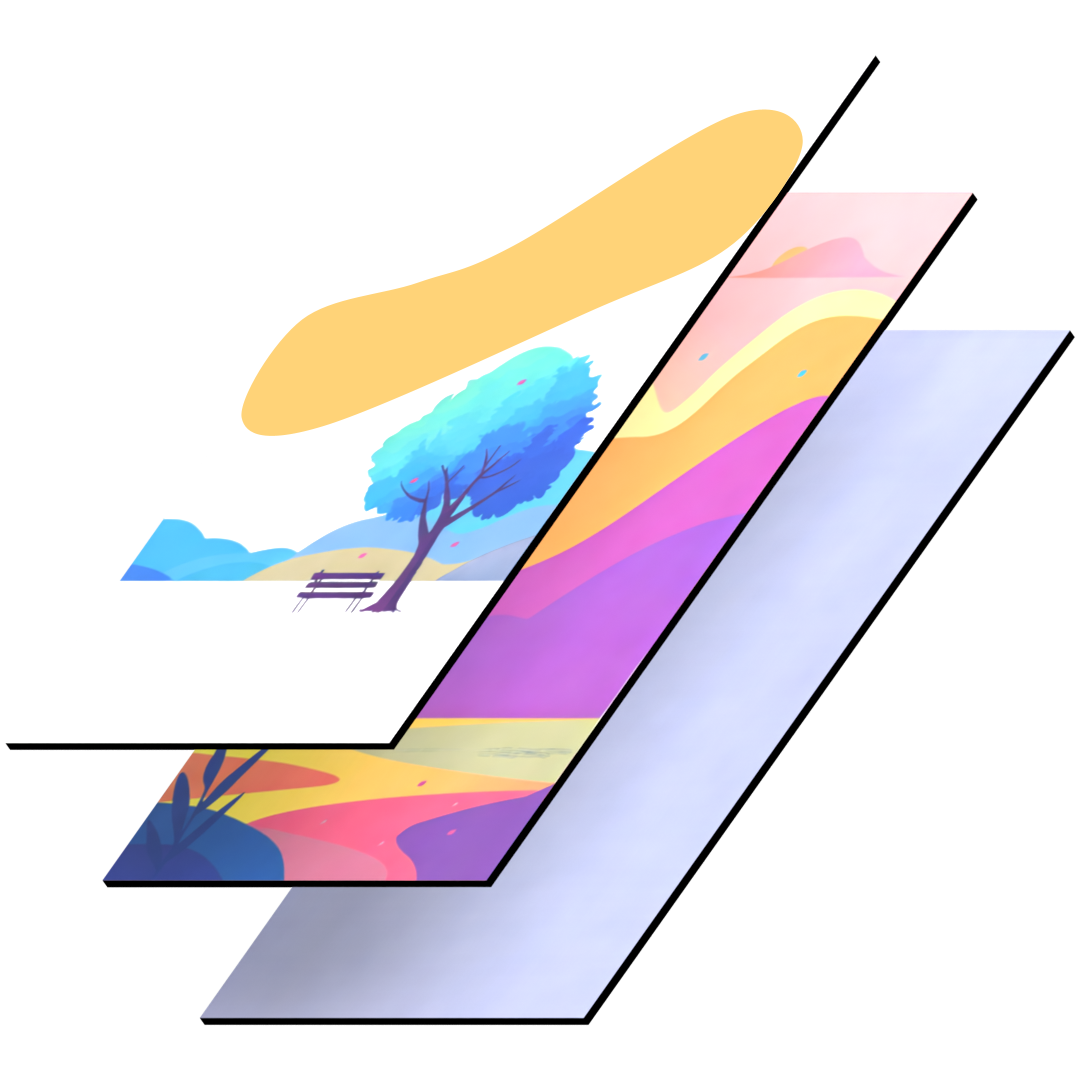}};
\node[anchor=center] (c3) at (0.30, -0.038)
    {\includegraphics[width=0.09\tw]{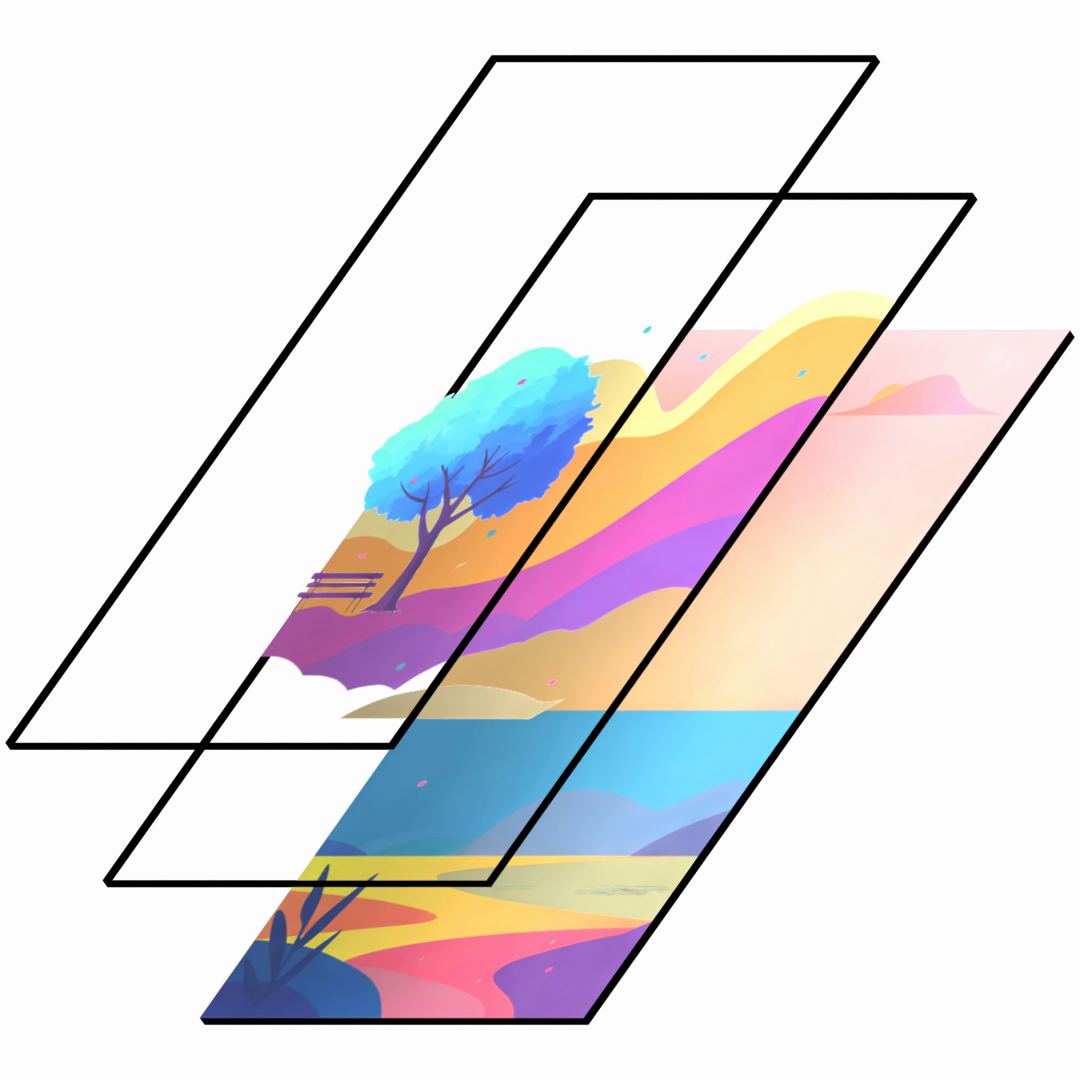}};
\node[anchor=center] (c4) at (0.40, -0.038)
    {\includegraphics[width=0.09\tw]{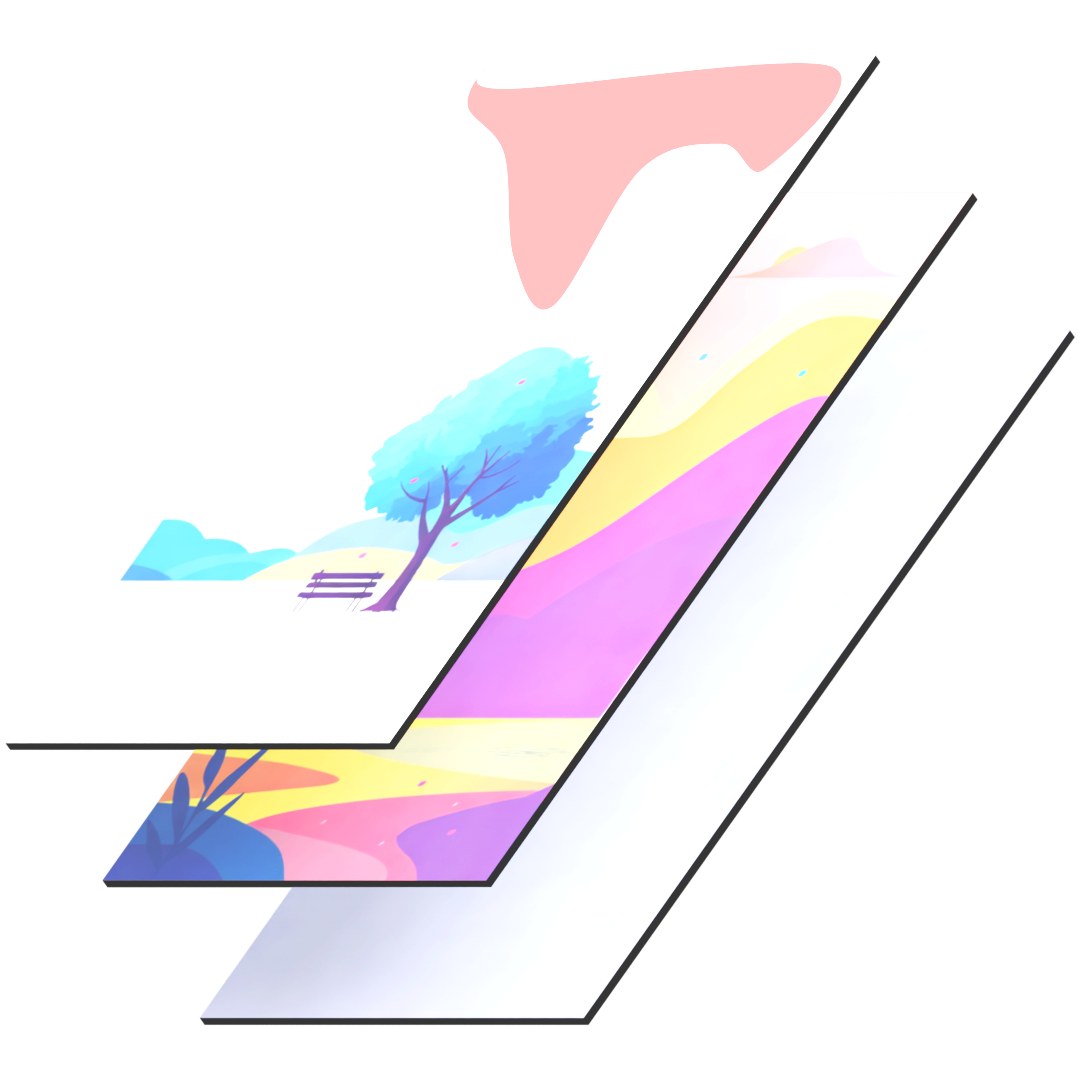}};

\node[draw=gray!60, dashed, rounded corners=4pt, line width=1pt,
      inner sep=0.007\tw, fit=(c1)(c2)(c3)(c4)] (cbox) {};

\coordinate (arr2-start) at ($(cbox.east)+(0.006,0)$);
\coordinate (arr2-end)   at (0.53, 0);
\draw[bigarrow] (arr2-start) -- (arr2-end);

\coordinate (arr2-mid) at ($(arr2-start)!0.5!(arr2-end)$);
\node[font=\large, anchor=north] at ($(arr2-mid)+(0,0.045)$)
    {\faRobot};
\node[font=\scriptsize, anchor=north, inner sep=1pt] at ($(arr2-mid)+(0,-0.015)$)
    {VLM};

\node[font=\footnotesize\bfseries, anchor=west] (s1) at (0.54, 0.014) {0.7};
\node[font=\footnotesize\bfseries, anchor=west] (s2) at (0.58, 0.014) {0.8};
\node[font=\footnotesize\bfseries, anchor=west] (s3) at (0.54, -0.006) {0.9};
\node[font=\footnotesize\bfseries, anchor=west] (s4) at (0.58, -0.006) {0.7};

\node[draw=gray!80, dashed, rounded corners=2pt, line width=0.8pt,
      inner sep=4pt, fit=(s1)(s2)(s3)(s4)] (sbox) {};
\node[font=\scriptsize, anchor=north, inner sep=1pt] at ($(arr2-mid)+(0.07,-0.03)$)
    {Scores};

\draw[bigarrow] ($(sbox.east)+(0.02, 0)$) -- ++(0.03, 0);

\node[anchor=center] (lora) at (0.80, 0)
    {\includegraphics[width=0.22\tw]{figures/teaser/tikz_imgs/candy_lora_white.png}};

\coordinate (grpo-up)   at ($(sbox.north)+(0, 0.01)$);   
\coordinate (grpo-top-r) at ($(sbox.north)+(0, 0.08)$);   
\coordinate (grpo-top-l) at (grpo-land |- grpo-top-r);     
\draw[line width=1.5pt, densely dashed, gray!80, rounded corners=4pt]
    (grpo-up) -- (grpo-top-r);
\draw[line width=1.5pt, densely dashed, gray!80, rounded corners=4pt]
    (grpo-top-r) -- (grpo-top-l);
\draw[line width=1.5pt, densely dashed, gray!80, ->, rounded corners=4pt]
    (grpo-top-l) -- (grpo-land);

\node[font=\bfseries\small, text=gray!80, fill=white, inner sep=2pt]
    at ($(grpo-top-r)!0.5!(grpo-top-l)+(0,0)$) {Flow-GRPO};

\node[lbl, anchor=north] at (0.07, -0.115)  {Input Image};
\node[lbl, anchor=north, text=gray!80] at (0.35, -0.09)  {Candidates};
\node[lbl, anchor=north] at (0.80, -0.1)  {Optimized Layers};

\draw[gray!30, line width=0.5pt] (-0.01, -0.145) -- (0.93, -0.145);

\def\brow{-0.26}

\node[anchor=center] (f_src) at (0.07, \brow)
    {\includegraphics[width=0.15\tw]{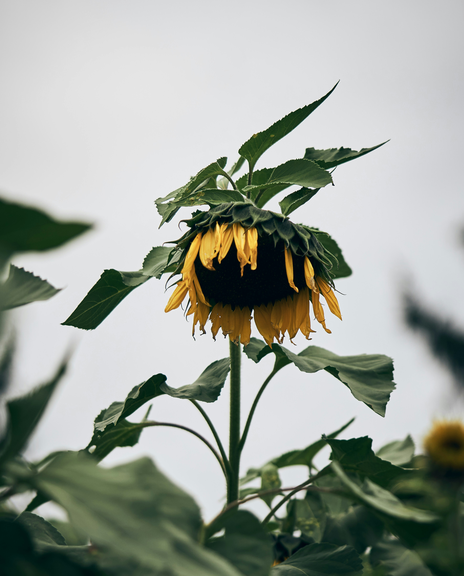}};

\draw[bigarrow] ($(f_src.east)+(0.008,0)$) -- ++(0.03, 0);

\node[anchor=center] (f_lora) at (0.33, \brow)
    {\includegraphics[width=0.22\tw]{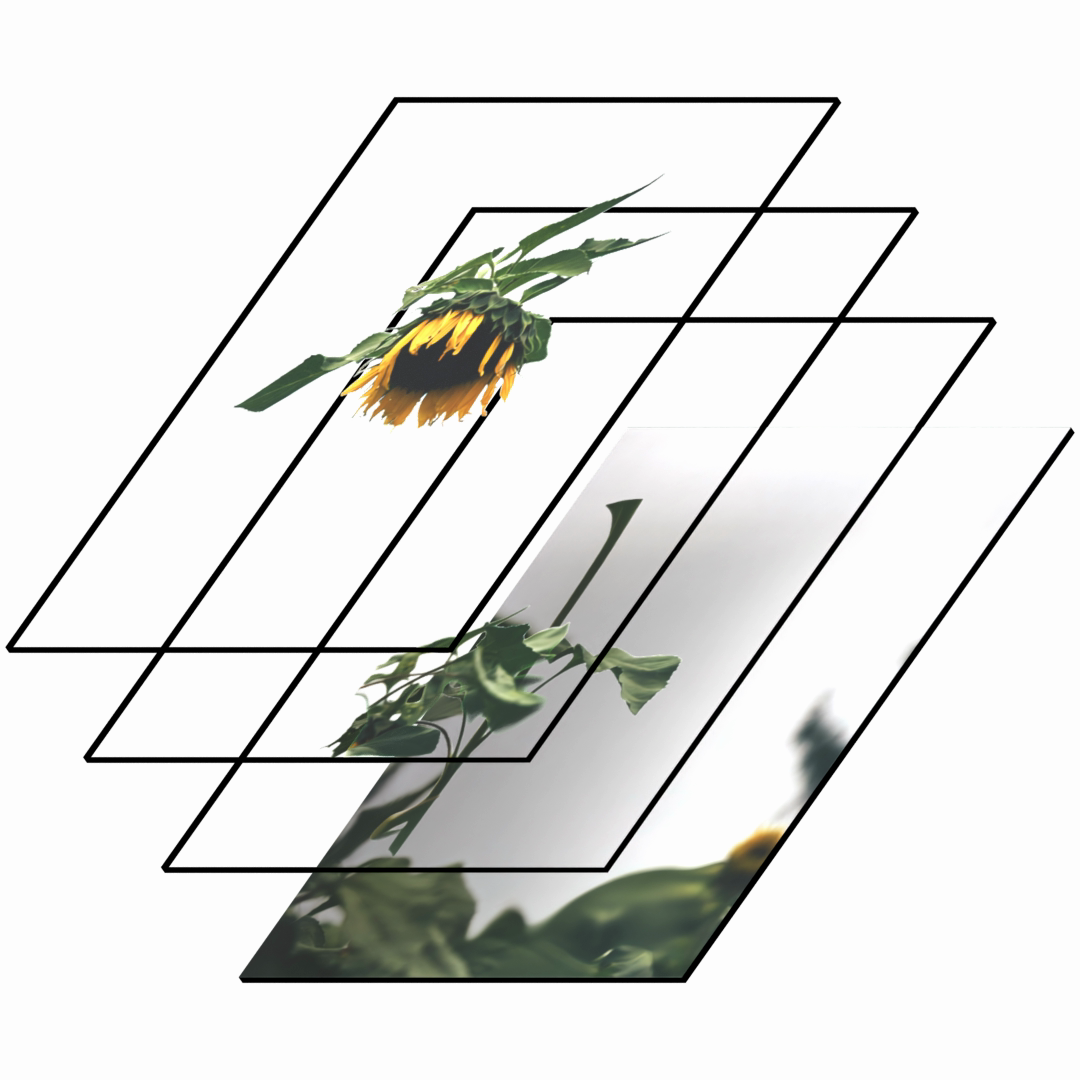}};

\draw[gray!30, line width=0.5pt] (0.47, \brow+0.085) -- (0.47, \brow-0.085);

\node[anchor=center] (r_src) at (0.57, \brow)
    {\includegraphics[width=0.12\tw]{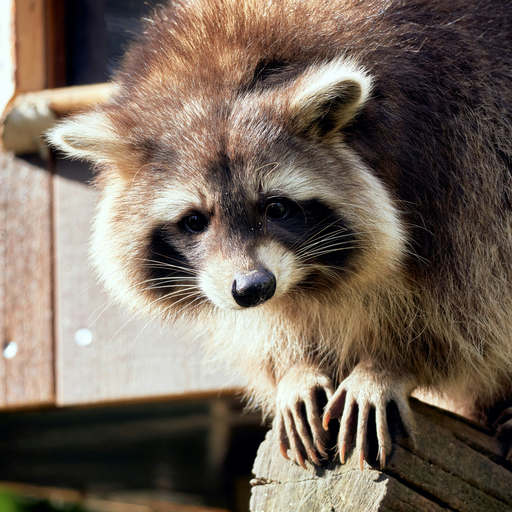}};

\draw[bigarrow] ($(r_src.east)+(0.008,0)$) -- ++(0.03, 0);

\node[anchor=center] (r_lora) at (0.80, \brow)
    {\includegraphics[width=0.22\tw]{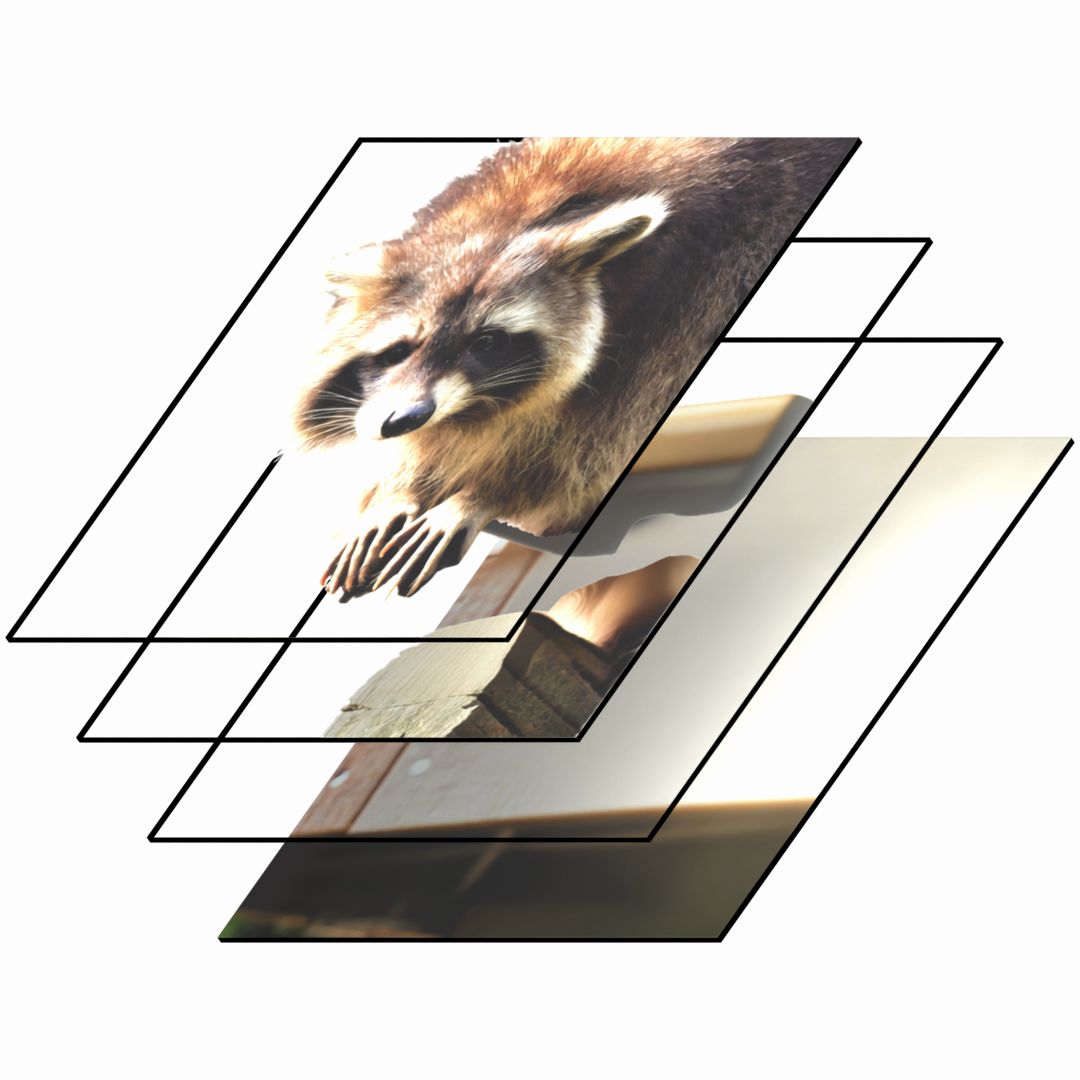}};

\end{tikzpicture} 
    }
    \vspace{-0.5em}
    \titlecaption{\method{}}{We finetune a layer decomposition model using Flow-GRPO and a VLM judge, and improve layerization without relying on paired data. The resulting layers have improved consistency, separation and handle in-painting of occluded areas better.}
    \label{fig:teaser}
\end{center}

\raggedbottom

\begin{abstract}
We present Stable-Layers, a reinforcement learning framework that eliminates the
need for paired supervision by fine-tuning a pretrained layer decomposition model
using only feedback from a vision-language model (VLM). Starting from
Qwen-Image-Layered, we apply Flow-GRPO with LoRA adaptation, sampling multiple
candidate decompositions per image, scoring them with a VLM, and optimising the
policy from group-relative advantages. The key challenge lies in designing a
reliable reward signal: VLMs scoring samples in isolation tend to compress their
judgements into a narrow band, leaving GRPO with little within-group variance
to learn from. We address this with a two-stage evaluation pipeline that pairs
structured per-sample scoring across five edit-centric criteria with a grid-based
calibration step in which the VLM re-scores all candidates side-by-side. Trained
entirely on unlabelled images, Stable-Layers produces decompositions with
stronger layer separation, fewer blank or artifact-heavy layers, and lower
per-layer reconstruction error on the Crello dataset compared to the base model.
\end{abstract}

\section{Introduction}
\label{sec:intro}

Image layer decomposition—separating an image into a small set of editable RGBA layers whose composition reconstructs the original (\Cref{fig:teaser})—is a fundamental primitive for professional editing and compositing~\citep{yin2025qwen}. While the task is easy to define, it is difficult to supervise: a single image admits many plausible decompositions, and quality is ultimately determined by downstream usability, including semantic separation, clean alpha mattes, minimal redundancy, and faithful handling of occluded content, rather than similarity to any single target decomposition. Existing methods circumvent this ambiguity using synthetic layered datasets~\citep{yin2025qwen,huang2024layerdiff,kang2025layeringdiff}, but such supervision imposes an inherent limitation: when multiple decompositions are equally valid, regression toward a single target penalizes alternative solutions. We address this limitation through a post-hoc reinforcement learning refinement stage that optimizes directly for perceived decomposition quality using a vision-language model (VLM) as the sole source of supervision~\citep{chen2024mjbench}.

However, applying VLM-as-judge feedback to layer decomposition introduces a reward design challenge not encountered by standard approaches. Decomposition quality is inherently multi-dimensional, spanning semantic disentanglement, alpha cleanliness, inpainting plausibility, feature allocation, and content validity, with strong correlations across these axes: candidates within a sampled group are often simultaneously good or bad along most criteria. Existing scalar VLM rewards~\citep{xu2026visionreward} collapse these dimensions into a single compressed signal, while pairwise preference approaches~\citep{wallace2024diffusiondpo} scale quadratically with group size and lose absolute calibration. Moreover, naive rubric scoring produces low within-group variance when candidates are visually similar, weakening the learning signal for GRPO. To address this, we introduce a two-phase evaluation protocol. In Phase~1, the VLM performs structured criterion-wise scoring using explicit rubric anchors for each quality dimension. In Phase~2, the candidate group is jointly re-evaluated on a labelled comparison grid to obtain finer relative calibration between similar decompositions. The two phases serve complementary roles: absolute scoring effectively captures categorical failures, while relative comparison sharpens discrimination between perceptually close candidates.

A second challenge is optimization stability. GRPO-Guard's RatioNorm~\citep{grpoguard2025} uses a spatial mean of per-element log-probabilities, but \modelname{} packs $N$ RGBA layers into a single latent sequence, inflating effective dimensionality $D$ by ${\sim}5\times$ and suppressing per-step log-ratio standard deviation as $1/\sqrt{D}$. To address this, we introduce a simple sum-and-rescale modification that restores $\mathcal{O}(1)$ ratio magnitudes while remaining broadly applicable to flow-matching RL settings with sequence-packed latents.

We instantiate our framework on \modelname{}~\citep{yin2025qwen} using LoRA adaptation~\citep{hu2021lora}, training entirely on Fine-T2I~\citep{ma2026finet2i} images without any layer annotations. Our contributions are: \textbf{(i)}~a two-phase VLM reward protocol that alleviates score compression in within-group reinforcement learning; \textbf{(ii)}~a RatioNorm reformulation tailored to packed latent representations in flow-matching RL; and \textbf{(iii)}~substantially improved layer decompositions over the \modelname{} baseline, yielding better semantic separation, cleaner layers, and lower per-layer reconstruction error on the Crello dataset \cite{yamaguchi2021canvasvae} and held-out evaluations.

In summary, \method{} serves as a general recipe for training 
edit-oriented generators using \emph{judge feedback} instead of targets.
The specific optimization machinery is a tool to achieve this goal: convert VLM judgments over sampled candidates into learning signals that directly enhance editability.

\section{Related Work}
\label{sec:related}
 
\inlinesection{RL and reward modelling for visual generation.} %
DDPO~\citep{black2024ddpo} and DPOK~\citep{dpok2023} first cast diffusion
sampling as a multi-step MDP and applied policy gradients to optimise
non-differentiable rewards; gradient-based alternatives such as
DRaFT~\citep{draft2024} and AlignProp~\citep{prabhudesai2023alignprop}
backpropagate through the sampling chain when the reward is differentiable,
while Diffusion-DPO~\citep{wallace2024diffusiondpo} sidesteps online rollouts
by optimising a preference-based objective on static comparison data.
For flow-matching specifically, Flow-GRPO~\citep{flowgrpo2025} introduces a
marginal-preserving SDE that yields tractable log-probabilities for
GRPO-style~\citep{shao2024deepseekmath} clipped objectives,
DanceGRPO~\citep{dancegrpo2025} validates group-relative updates at scale,
and GRPO-Guard~\citep{grpoguard2025} stabilises the importance ratio via
normalisation and gradient reweighting.
Reward signals for these methods range from learned scalar rewards
(ImageReward~\citep{xu2023imagereward}, HPS\,v2~\citep{wu2023hpsv2}) to
VLM-derived rewards: scalar rewards from pretrained
encoders~\citep{xu2026visionreward}, pairwise
preferences for DPO-style
training~\citep{zhang2025direct}, and self-improving
VLM critics~\citep{li2025self,evoquality2025}.
TOPReward~\citep{topreward2026} extracts token-completion logits to bypass
the brittleness of text-generated numeric scores, and
MJ-Bench~\citep{chen2024mjbench} studies VLM reliability as a judge.
Our training loop follows Flow-GRPO with GRPO-Guard stabilisation, but
replaces scalar or pairwise reward interfaces with structured multi-criteria
VLM judgements (alpha cleanliness, semantic separation, content validity)
that enable richer credit assignment over layer stacks.
 
\inlinesection{Image layer decomposition and generation.} %
Recovering or synthesizing layered image representations has been approached
from multiple directions.
Text-to-layer generation methods, including
LayerDiff~\citep{huang2024layerdiff},
DreamLayer~\citep{huang2025dreamlayer},
LayerFusion~\citep{dalva2024layerfusion},
PSDiffusion~\citep{huang2025psdiffusion}, and
LayeringDiff~\citep{kang2025layeringdiff}, produce multi-layer raster outputs
via inter-layer attention, harmonized decoding, or generate-then-disassemble
pipelines.
LayerDiffuse~\citep{zhang2024layerdiffuse} encodes alpha-channel transparency
in the latent manifold of a pretrained diffusion model for direct RGBA
generation.
On the decomposition side, LayerDecomp~\citep{yang2025generative} and
LASAGNA~\citep{yang2026lasagna} separate foreground and background while
preserving visual effects;
Referring Layer Decomposition~\citep{reflayer2026} conditions on user prompts;
CLD~\citep{cld2025} introduces fine-grained controllable multi-layer
separation; and Chen~\etal~\citep{chen2025inpainting2layer} repurpose
inpainting models for layer recovery.
Domain-specific methods target graphic designs~\citep{suzuki2025layerd},
anime characters~\citep{lin2026seethrough}, and illustration production
workflows~\citep{zhang2026workflow}.
Most closely related to our work,
Qwen-Image-Layered~\citep{yin2025qwen} is an end-to-end diffusion model that
decomposes a single RGB image into a variable number of RGBA layers using an
RGBA-VAE, a Variable Layers Decomposition MMDiT, and multi-stage supervised
training on Photoshop PSD data.
We take Qwen-Image-Layered as our base model and show that GRPO-based
reinforcement learning with VLM-as-judge rewards can further improve
decomposition quality beyond what supervised training alone achieves.

\section{Background}
\label{sec:background}

Our method builds on three components: flow matching as the generative
framework, an SDE-augmented variant that exposes tractable per-step
log-probabilities, and GRPO for policy optimisation.

\inlinesection{Flow Matching and Rectified Flows} %
Rectified flow~\citep{liu2022flowa,liu2022flowb} interpolates
$x_t = (1-t)\,x_0 + t\,x_1$ between data $x_0 \sim \pi_{\mathrm{ref}}$ and
noise $x_1 \sim \mathcal{N}(0, I)$, and learns a velocity field
$\vtheta(x_t, t)$ parameterized by $\theta$ via the regression loss
$\mathcal{L}_{\mathrm{FM}} = \E_{t, x_0, x_1}\bigl[\| \vtheta(x_t, t) - (x_1 - x_0) \|^2 \bigr]$.
At inference, samples are produced by integrating the ODE
$x_{t - \Delta t} = x_t + \vtheta(x_t, t)\,\Delta t$. This deterministic
trajectory has no per-step randomness and thus no tractable
log-probability for policy gradients; the SDE formulation below
addresses this.

\inlinesection{SDE-Augmented Flow Matching (Flow-GRPO)} %
Flow-GRPO~\citep{flowgrpo2025} augments the deterministic ODE with a
stochastic differential equation that preserves the learned marginals,
enabling tractable log-probability computation for RL. With diffusion
coefficient $\sigma_t = a\sqrt{t/(1-t)}$ (we use $a{=}0.7$, within the
$[0.7, 0.9]$ range recommended by \citet{flowgrpo2025}), the SDE
transition for a step from $t$ to $t-\Delta t$ is:
\begin{align}
  \mu_{t \to t-\Delta t} &= x_t \left(1 - \frac{\sigma_t^2}{2t} \Delta t \right)
    - \vtheta(x_t, t) \left(1 + \frac{\sigma_t^2 (1-t)}{2t}\right) \Delta t, \\
  x_{t - \Delta t} &= \mu_{t \to t-\Delta t} + \sigma_t \sqrt{\Delta t}\; \epsilon,
    \quad \epsilon \sim \mathcal{N}(0, I).
  \label{eq:sde-step}
\end{align}

\inlinesection{Group Relative Policy Optimization} %
For a group of $G$ samples $\{x^{(g)}\}_{g=1}^G$ from the same condition,
GRPO~\citep{shao2024deepseekmath} computes within-group advantage $\hat{A}^{(g)}$ from each sample's reward $r^{(g)}$ as:
\begin{equation}
  \hat{A}^{(g)} = \frac{r^{(g)} - \bar{r}}{\sigma_r + \nu},
  \label{eq:grpo-advantage}
\end{equation}
where $\bar{r}$ and $\sigma_r$ are the within-group mean and standard
deviation of rewards and $\nu{=}10^{-4}$ is a small constant for numerical
stability. GRPO optimises a clipped surrogate
$\mathcal{L}_{\mathrm{GRPO}} = -\E_g[\min(\rho_g \hat{A}^{(g)},\,
\mathrm{clip}(\rho_g, 1{-}\epsilon_c, 1{+}\epsilon_c)\hat{A}^{(g)})]
+ \beta\,\KL[\pi_\theta \| \pi_{\mathrm{ref}}]$,
where $\rho_g = \exp(\log\pi_\theta - \log\pi_{\theta_\mathrm{old}})$
is the importance ratio.

\inlinesection{Stabilised Ratio Clipping (GRPO-Guard)}
\label{sec:bg-grpoguard}
\citet{grpoguard2025} observe that in flow-matching models the importance
ratio $\rho_g$ exhibits a systematic leftward shift (mean below~$1$) with
timestep-dependent variance, preventing the clipped surrogate from
constraining overconfident positive-advantage updates.
GRPO-Guard addresses this with two corrections:
(i)~\emph{RatioNorm}, which standardizes $\log \rho_g$ per denoising
step so that the ratio distribution is centered near~$1$ with uniform
variance across steps, restoring effective clipping; and
(ii)~\emph{gradient reweighting} by $\delta = 1/\Delta t$, which
equalizes per-step gradient magnitudes and prevents low-noise timesteps
from dominating the update, allowing the KL term to be omitted.

\section{Method}
\label{sec:method}

\method{} fine-tunes a pretrained layer decomposition model using only
unlabeled images and post-hoc VLM judgements as supervision---no layer
annotations, paired examples, or synthetic decomposition targets are
required. 
We adopt Flow-GRPO's three-phase training loop
(\Cref{fig:pipeline}): each step generates $G$ candidate
decompositions via SDE sampling, scores them with the VLM reward
protocol of \Cref{sec:method-reward}, and replays the stored
trajectories to compute GRPO updates. The reward design
(\Cref{sec:method-reward}), data strategy (\Cref{sec:method-data}),
and adaptation to an editing model
(\Cref{sec:method-architecture}) are our contributions; the SDE
formulation and GRPO objective follow Flow-GRPO directly.

\begin{figure}[t]
  \centering
  \resizebox{\textwidth}{!}{
    \begin{tikzpicture}[
    >=Stealth,
    mainbox/.style={
        rounded corners=5pt,
        line width=1.2pt,
        minimum width=52mm,
        minimum height=32mm,
        align=center,
        font=\large,
        inner sep=6pt,
    },
    subbox/.style={
        rounded corners=3pt,
        draw=detailPurple,
        fill=detailPurpleBg,
        line width=0.8pt,
        minimum width=44mm,
        align=center,
        font=\normalsize,
        inner sep=5pt,
    },
    inputbox/.style={
        rounded corners=2pt,
        draw=arrowGray,
        fill=white,
        line width=0.7pt,
        align=center,
        font=\small,
        inner sep=4pt,
    },
    storedbox/.style={
        rounded corners=2pt,
        draw=arrowGray!60,
        fill=white,
        line width=0.7pt,
        align=center,
        font=\normalsize,
        inner sep=5pt,
    },
    secfont/.style={font=\small\itshape, text=secRef},
    arrlab/.style={font=\normalsize, text=arrowGray, fill=white, inner sep=2pt},
    mainlabel/.style={font=\Large\bfseries},
    bigarrow/.style={->, line width=1.6pt, arrowGray},
]


\node[mainbox, draw=genBlue, fill=genBlueBg] (gen) {
    \textcolor{genBlue}{\Large\bfseries 1.\;Generate}\\[5pt]
    Sample $G$ decompositions\\
    via SDE flow matching\\
    (no gradient)
};

\node[mainbox, draw=scoreRed, fill=scoreRedBg, right=16mm of gen] (score) {
    \textcolor{scoreRed}{\Large\bfseries 2.\;Score}\\[5pt]
    VLM reward model\\
    two-phase scoring\\
    $r^{(1)},\,\ldots,\,r^{(G)}$
};

\node[mainbox, draw=trainGreen, fill=trainGreenBg, right=16mm of score] (train) {
    \textcolor{trainGreen}{\Large\bfseries 3.\;Train}\\[5pt]
    Replay trajectories\\
    GRPO loss\\
    update LoRA parameters
};

\node[inputbox, left=5mm of gen] (input) {
    unlabeled images\\
    {\footnotesize Fine-T2I}
};

\draw[bigarrow] (input.east) -- (gen.west);
\draw[bigarrow] (gen.east) -- node[arrlab, above, yshift=1mm] {layers} (score.west);
\draw[bigarrow] (score.east) -- node[arrlab, above, yshift=1mm] {rewards} (train.west);

\draw[->, line width=1.4pt, loopGreen, rounded corners=6pt]
    (train.north) -- ++(0, 5mm)
    -| node[arrlab, above, pos=0.25, text=loopGreen, font=\normalsize\itshape] {repeat}
    (gen.north);

\node[secfont, below=1mm of gen.south east, anchor=north east, xshift=2mm]
    {\S\,3.1, \S\,3.4};
\node[secfont, below=1mm of score.south west, anchor=north west, xshift=-2mm]
    {\S\,3.2};
\node[secfont, below=1mm of train.south west, anchor=north west, xshift=-2mm]
    {\S\,3.1, \S\,3.4};
\node[secfont, below=1mm of input.south]
    {\S\,3.3};

\node[storedbox, below=6mm of gen] (stored) {
    stored:\; $\tau^{(g)},\;\log \pi_{\theta_{\mathrm{old}}}$
};
\draw[->, line width=0.9pt, arrowGray!70] (gen.south) -- (stored.north);

\node[subbox, below=6mm of score] (phase1) {
    \textbf{Phase 1: structured rubric}\\[2pt]
    {\small 5 criteria, 0--5 each}
};
\draw[->, line width=0.9pt, arrowGray!70] (score.south) -- (phase1.north);

\node[subbox, below=6mm of phase1] (phase2) {
    \textbf{Phase 2: grid calibration}\\[2pt]
    {\small relative re-scoring}
};
\draw[->, line width=0.9pt, arrowGray!70] (phase1.south) -- (phase2.north);

\node[font=\normalsize\bfseries, text=detailPurple,
      right=0mm of phase2.north east, anchor=south west]
    (vlmlabel) {VLM Judge};

\node[subbox, draw=trainGreen!70, fill=trainGreenBg,
      below=6mm of train] (lora) {
    \textbf{LoRA}\; ($r{=}16,\;\alpha{=}16$)\\[2pt]
    {\small attention + FF layers}
};
\draw[->, line width=0.9pt, arrowGray!70] (train.south) -- (lora.north);

\node[subbox, draw=trainGreen!70, fill=trainGreenBg,
      below=6mm of lora] (ppo) {
    $K{=}1$ \textbf{PPO epoch}\\[2pt]
    {\small clipped surrogate}
};
\draw[->, line width=0.9pt, arrowGray!70] (lora.south) -- (ppo.north);

\end{tikzpicture}
  }
  \titlecaption{\method{} training pipeline}{Sample $G$ candidates, score with the two-phase VLM reward, replay with GRPO updates to LoRA parameters.}
  \label{fig:pipeline}
  \vspace{-1.5em}
\end{figure}

\subsection{Model Architecture and Adaptation}
\label{sec:method-architecture}

The base model we use is \modelname{}~\citep{yin2025qwen}, a flow-matching
transformer~\citep{liu2022flowa, liu2022flowb} that generates
$N$-layer RGBA decompositions conditioned on an input image.
The architecture comprises a 3D variational autoencoder (VAE) that encodes
4-channel RGBA frames with $8{\times}$ spatial compression into a 16-channel
latent space, a sequence-based transformer operating on $2{\times}2$
patch-packed latents (yielding token dimension $16 \times 4 = 64$), and a
text encoder for prompt conditioning.
The condition image is encoded through the same VAE pipeline and concatenated
along the sequence dimension of the transformer input, with per-frame spatial
metadata provided to the attention mechanism to distinguish generated layer
tokens from conditioning tokens.

We apply Low-Rank Adaptation~\citep{hu2021lora} with rank $r{=}16$ and
$\alpha{=}16$ to all attention projection layers and feed-forward layers,
keeping all other parameters frozen.

\subsection{VLM Reward Design}
\label{sec:method-reward}
The central challenge in applying RL to layer decomposition is defining a
reward signal that captures the multi-dimensional notion of decomposition
quality without requiring ground-truth targets.
We address this with a two-phase VLM scoring pipeline, designed to avoid
specific failure modes of the base model while maintaining inter-group
discrimination. Each criterion in our rubric is anchored by explicit
descriptions of qualifying high- and low-score conditions, which we found
necessary for consistent VLM judgements; the full rubric is provided in
\Cref{app:reward-prompt}.

\subsubsection{Image Presentation for the VLM Judge}
\label{sec:method-reward-presentation}

VLMs are trained predominantly on RGB and cannot meaningfully
interpret raw alpha channels, so we composite each layer onto a solid
white background before presenting it: transparent regions render as
white, and the VLM assesses alpha quality by observing where content
transitions to the white background. Each sample is presented as the
RGB composite alongside $N$ white-background layer images at
$320 \times 320$ for Phase~1.

\subsubsection{Phase 1: Structured Individual Scoring}
\label{sec:method-reward-individual}

Each generated sample is sent independently to the VLM, which
evaluates five criteria on a $0$--$5$ integer scale, each anchored by
explicit descriptions of the worst-case ($0$) and best-case ($5$)
visual conditions. The criteria are: \textbf{semantic separation}
(each foreground layer isolates one distinct object), \textbf{alpha
cleanliness} (foreground masks are crisp and binary-like),
\textbf{background inpainting} (layer~0 is a plausible scene
completion), \textbf{feature distribution} (content is spread across
layers rather than concentrated), and \textbf{content validity}
(layers are not blank or noise-only). The five scores sum to a total
in $[0, 25]$ and are normalised to $[0, 1]$. The full prompt with
per-criterion anchor descriptions is in \Cref{app:reward-prompt}.

\begin{figure}[t]
  \begin{minipage}[c]{0.45\linewidth}
    \resizebox{\linewidth}{!}{
    \begin{tikzpicture}[
    >={Stealth[length=2.8mm,width=2.4mm]},
    smallbox/.style={draw=boxedge, fill=boxfill, rounded corners=1pt,
                     minimum size=5mm, line width=0.3pt},
    innerbox/.style={draw=boxedge, fill=boxfill, rounded corners=1pt,
                     minimum size=3mm, line width=0.2pt},
    bigbox/.style={draw=boxedge, fill=boxfill, rounded corners=2.5pt,
                   minimum size=16mm, line width=0.5pt},
    judgebox/.style={draw=vlmedge, fill=vlmfill, rounded corners=2pt,
                     text=vlmtext, font=\bfseries, inner sep=2pt, line width=0.5pt},
    arr/.style={->, draw=arrowgrey, line width=0.9pt},
    num/.style={font=\ttfamily\small},
    every node/.style={font=\sffamily}
]

\def\xL{0}\def\xLn{0.65}    
\def\xM{4.3}                
\def\xR{8.0}\def\xRn{8.65}  
\def\yA{3.0}\def\yB{2.5}\def\yC{2.0}\def\yD{1.5}
\def\yMid{2.2}            

\node[font=\bfseries] at (\xL, 3.85) {Phase 1};
\node[smallbox] (a1) at (\xL,\yA) {};
\node[smallbox] (a2) at (\xL,\yB) {};
\node[smallbox] (a3) at (\xL,\yC) {};
\node[smallbox] (a4) at (\xL,\yD) {};
\node[num, text=numgrey] at (\xLn,\yA) {0.72};
\node[num, text=numgrey] at (\xLn,\yB) {0.74};
\node[num, text=numgrey] at (\xLn,\yC) {0.71};
\node[num, text=numgrey] at (\xLn,\yD) {0.73};

\node[bigbox] (vlm) at (\xM,\yMid) {};
\node[innerbox] at ([xshift=-5mm, yshift= 5mm]vlm.center) {};
\node[innerbox] at ([xshift= 5mm, yshift= 5mm]vlm.center) {};
\node[innerbox] at ([xshift=-5mm, yshift=-5mm]vlm.center) {};
\node[innerbox] at ([xshift= 5mm, yshift=-5mm]vlm.center) {};

\node[font=\bfseries] at (\xR, 3.85) {Phase 2};
\node[smallbox] (b1) at (\xR,\yA) {};
\node[smallbox] (b2) at (\xR,\yB) {};
\node[smallbox] (b3) at (\xR,\yC) {};
\node[smallbox] (b4) at (\xR,\yD) {};
\node[num, text=posgreen] at (\xRn,\yA) {0.82};
\node[num, text=posgreen] at (\xRn,\yB) {0.91};
\node[num, text=negred]   at (\xRn,\yC) {0.38};
\node[num, text=negred]   at (\xRn,\yD) {0.45};

\draw[arr] (1.7, \yMid) -- (vlm.west);
\draw[arr] (vlm.east) -- (6.9, \yMid);

\node[judgebox] (judge) at (\xM, 3.75) {VLM Judge};
\draw[arr] (judge.south) -- (vlm.north);

\end{tikzpicture}
  }
  \end{minipage}%
  \hfill
  \begin{minipage}[c]{0.46\linewidth}
    \titlecaption{Phase~2 grid calibration}{We re-score candidates relative
      to each other, spreading compressed Phase~1 scores and
      restoring within-group variance for GRPO advantage
      normalisation.}
    \label{fig:calibration}
  \end{minipage}
  \vspace{-1em}
\end{figure}

\subsubsection{Phase 2: Relative Grid Calibration}
\label{sec:method-reward-calibration}

To sharpen within-group discrimination when Phase~1 scores ($r_{\mathrm{ind}}^{(g)}$) compress,
we tile all $G$ composites into a labelled comparison grid at a resolution of 256x256
(\Cref{fig:calibration}) and ask the VLM to re-score each sample
relative to the others ($r_{\mathrm{cal}}^{(g)}$), given the Phase~1 scores as context.
Construction details (cell size, label format, full prompt) are in
\Cref{app:reward-prompt}.

We use the calibrated score directly: $r^{(g)} = r_{\mathrm{cal}}^{(g)}$. Phase~2 conditions on the Phase~1 scores (\Cref{app:reward-prompt}). The no-calibration baseline (\Cref{sec:ablation-calibration}) substitutes $r_{\mathrm{ind}}^{(g)}$.

\subsection{Training Data: Judge-Only Supervision without Synthetic Targets}
\label{sec:method-data}

The primary source is Fine-T2I~\citep{ma2026finet2i}, an aesthetically
filtered subset of photographs and artworks.
All images are resized to $640 \times 640$ and normalised to $[-1, 1]$;
The images are shuffled at each epoch.
At each training step, the number of output layers is sampled uniformly from
$[\text{min\_layers}, \text{max\_layers}]$ (typically $[2, 5]$), exposing the
model to variable decomposition complexity throughout training.
While the base \modelname{} model supports decompositions of 20 layers,
the memory and compute cost of GRPO scales with both the group size $G$ and
the number of layers per sample (each additional layer adds tokens to the
transformer's sequence and a separate VLM scoring pass), so we restrict
training to at most five layers per sample. The
trained LoRA can be applied with the full range of possible output layer numbers during inference.

\subsection{GRPO Training with Trajectory Replay}
\label{sec:method-training}

We follow Flow-GRPO's trajectory replay procedure with one
modification to GRPO-Guard's RatioNorm: because \modelname{} packs
multiple RGBA layers into a single high-dimensional latent sequence
at $640\times 640$, the standard spatial-mean log-ratio collapses
toward zero. We instead sum log-probabilities over spatial dimensions and normalise by $\sqrt{D}$ before applying GRPO-Guard's per-step centring, preserving $\mathcal{O}(1)$ ratios while retaining RatioNorm's centring and variance-stabilisation properties. Full hyperparameters,
SDE step schedules, and CFG settings are in
\Cref{app:training-details}.

\section{Experimental Setup}
\label{sec:experiments}

The images from the training dataset are resized to $640 \times 640$ and normalised to $[-1, 1]$.
The number of output layers per step is sampled uniformly from $[2, 5]$
during training. For the Crello dataset evaluation (\Cref{tab:rgb-l1-crello}) we
restrict generation to $L \in \{2, 3, 4\}$ for direct comparison against
the base model on the same layer counts.

\subsection{Baselines}
\label{sec:exp-baselines}

We compare \method{} (full two-phase reward with grid calibration)
against two reference points:

\paragraph{Base model.}
The base \modelname{} checkpoint with no RL fine-tuning.
This establishes the baseline to compare to for the recipe.

\paragraph{Flow-GRPO without calibration.}
The same GRPO training pipeline with identical hyperparameters, LoRA
configuration, and data, but using only Phase~1 individual scoring as
the reward signal, no grid calibration phase).

\paragraph{Comparison with LayerD.}
We additionally compare against LayerD~\citep{suzuki2025layerd}, which
represents a different point in the design space: a method that
declines to decompose under uncertainty, frequently returning the
input largely intact as a single layer rather than producing a
multi-layer separation. This is a valid design choice with different
downstream implications than ours, and the comparison is included to
characterise the behavioural contrast. Setup details are in
\Cref{app:layerd-setup}.

We do not compare against supervised fine-tuning (SFT) with
reconstruction loss, as this requires paired ground-truth layer
decompositions that do not exist for natural images---the
supervision gap that motivates \method{}.

\section{Results}
\label{sec:results}

\subsection{Reward Progression}
\label{sec:results-reward}

\begin{figure}[!htbp]
  \vspace{-.75em}
  \centering
  \begin{minipage}[c]{0.4\linewidth}
    \includegraphics[width=\linewidth]{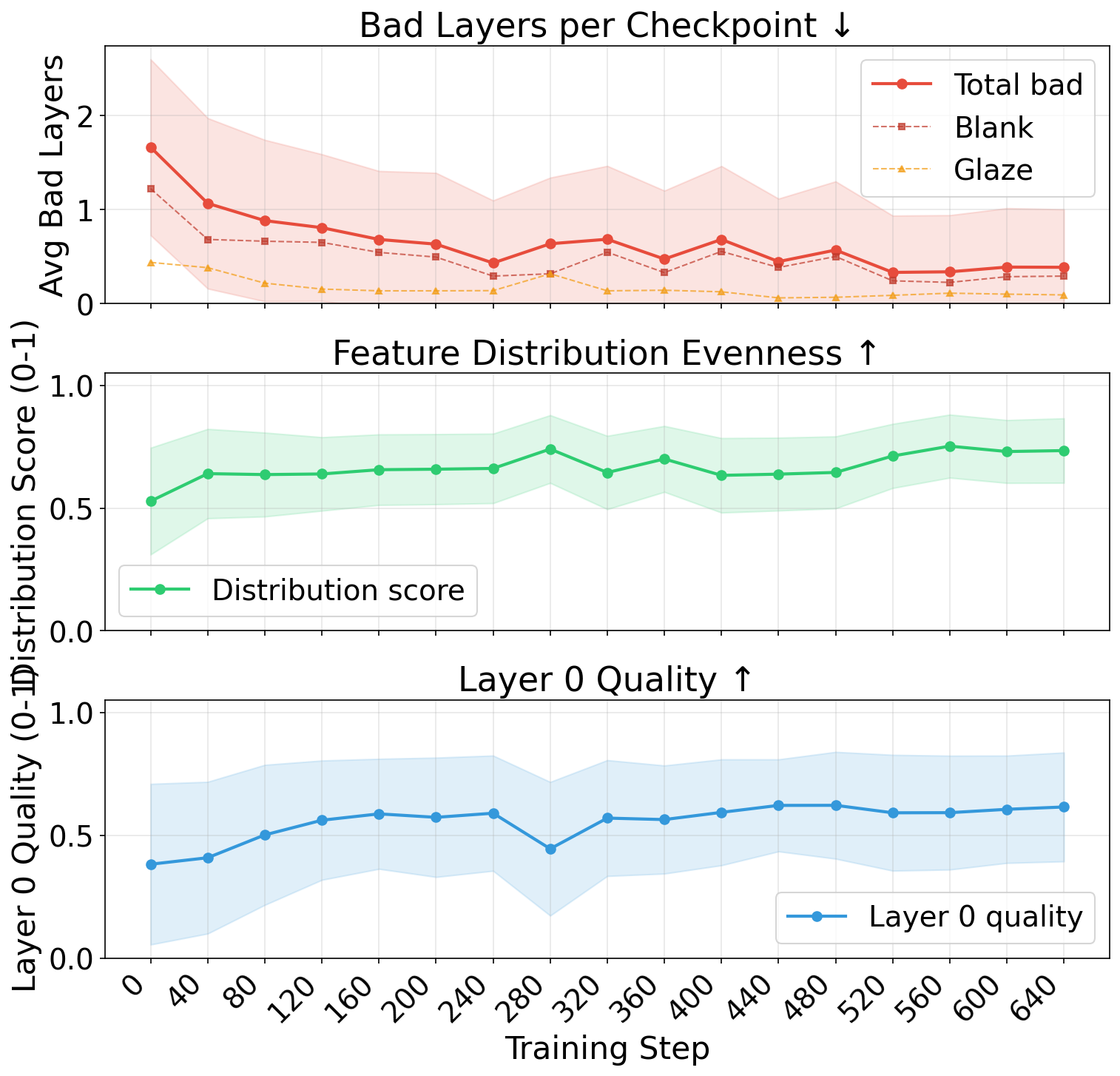}
  \end{minipage}%
  \hfill
  \begin{minipage}[c]{0.55\linewidth}
    \titlecaption{Held-out evaluation metrics}{Three automated metrics
      on 480 LAION-Aesthetics~\citep{schuhmann2022laion5b} images
      across training. \textbf{Top:} bad layers per decomposition
      (blank + glaze; lower is better) fall from ${\sim}1.65$ to
      ${\sim}0.4$. \textbf{Middle:} feature distribution evenness
      (higher is better) rises from ${\sim}0.53$ to ${\sim}0.73$.
      \textbf{Bottom:} layer~0 inpainting quality (higher is better)
      rises from ${\sim}0.38$ to ${\sim}0.62$. Bands show $\pm 1\sigma$
      across the set.}
    \label{fig:eval-metrics}
  \end{minipage}
  \vspace{-1em}
\end{figure}

The calibrated VLM reward rises from ${\sim}0.70$ to ${\sim}0.83$ over
the first ${\sim}100$ steps as the policy eliminates the worst
failure modes, then plateaus with high per-step variance for the
remainder of training (\Cref{fig:reward-progression}), even as
held-out evaluation metrics continue to improve
(\Cref{fig:eval-metrics}). This plateau is expected under GRPO:
because advantages are normalised \emph{within} each group
(\Cref{eq:grpo-advantage}), learning requires only sufficient
within-group variance to distinguish better candidates from worse,
not rising absolute scores---once the coarse failure modes are
resolved, all candidates in a group tend to improve together, keeping
the group mean roughly stationary while relative ranking continues to
provide gradient signal. The residual variance in the curve reflects
conditioning-image difficulty across the dataset more than policy
quality.

\subsection{Qualitative Results}
\label{sec:results-qualitative}

\begin{figure}[t]
  \centering
  \includegraphics[width=\linewidth]{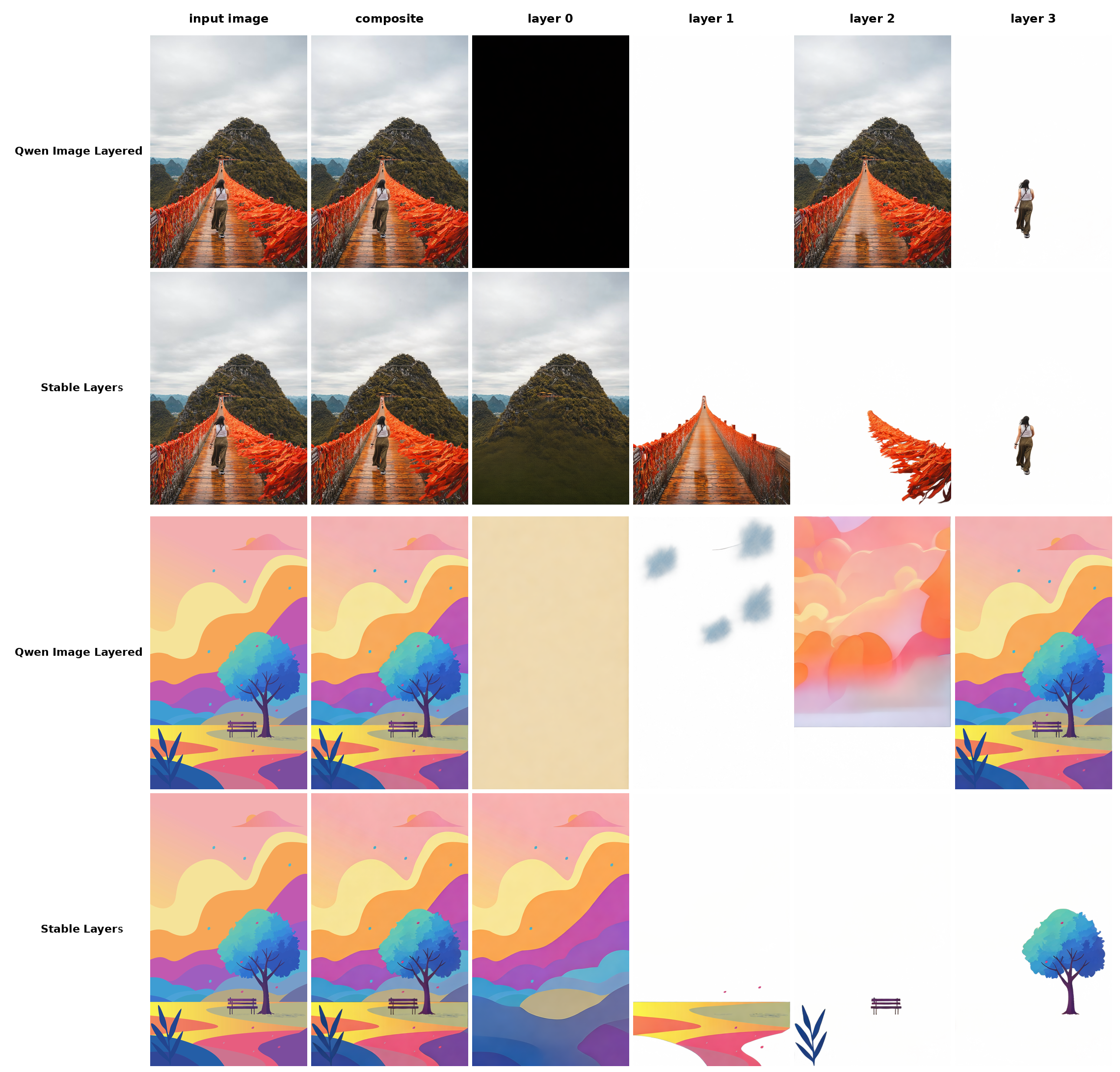}
  \titlecaption{Qualitative comparison on held-out images}{%
    Base model (\emph{Qwen Image Layered}, top of each pair) vs.\
    \method{}-fine-tuned model (\emph{Stable-Layers}, bottom).
    Columns show the input, the composite, and individual layers on
    white backgrounds. The fine-tuned model produces plausible
    background inpainting on layer~0 and isolates distinct semantic
    elements across foreground layers, where the base model leaves
    layer~0 degenerate and duplicates the composite across foreground
    slots.}
  \label{fig:qualitative}
\end{figure}

\Cref{fig:qualitative} compares decompositions from the base model
(\modelname{}) and the \method{} model on two held-out
inputs: a natural photograph (a person crossing a red rope bridge)
and a vector-style illustration (a tree and bench in a stylised
landscape). Two failure modes of the base model are visible across
both examples and addressed after fine-tuning. First, layer~0 is
degenerate---a fully black layer for the bridge scene and a flat
cream fill for the illustration, neither of which represents a usable
inpainted background. After fine-tuning, layer~0 contains a plausible
scene completion: the mountain and sky behind the bridge, and the
rolling hills and clouds of the illustration, demonstrating that the
VLM reward successfully penalises trivial inpainting. Second, the
base model duplicates near-complete copies of the input across
foreground layers (most clearly in the bridge example, where layer~2
is essentially the entire composite minus the person). The
fine-tuned model instead isolates distinct semantic elements onto
separate layers---the bridge deck, the rope railings, and the person
in the photograph; the foreground path, plants, bench, and tree in
the illustration---with cleaner alpha masks and less colour bleed
into transparent regions.

\vspace{-0.5em}

\subsection{Quantitative Evaluation}
\label{sec:results-quantitative}

We evaluate per-layer reconstruction quality on the
Crello dataset~\citep{yamaguchi2021canvasvae} test set using a metric adapted
from \citet{yin2025qwen}, with one modification: best-match assignment
in place of fixed-index comparison, since RL fine-tuning can reorder
layers without changing decomposition quality (full justification in
\Cref{app:crello-metric}).

\Cref{tab:rgb-l1-crello} reports per-layer RGB L1 against best-matched
ground-truth layers, stratified by output layer count. \method{} achieves
lower mean error than the base model across all layer counts. At the individual-layer level, the fine-tuned model reduces dominance of layer~0 (Pred\,0), consistent with the improved background inpainting seen in \Cref{fig:qualitative}. Small per-slot regressions (e.g.\ Pred\,1 at $L{=}2$ and Pred\,3 at $L{=}4$) reflect content reorganisation under best-match assignment: the fine-tuned model redistributes content across slots, so the slot that previously held the smallest residual (a near-empty layer in the base model) is now populated with real content and matches a different GT layer. Mean error is the relevant aggregate, and improves at every $L$.

\begin{table}[t]
    \caption{Per-layer RGB L1 vs Crello ground-truth test set layers 
        — each predicted layer is scored against its closest
        GT layer; lower is better). \emph{Mean} is the mean over all predicted
        layers; \emph{pred\,$k$} reports layer~$k$ individually. $m$ = number of
        matched Crello test set images. Bold marks the best \emph{Mean} within
        each $L$ group.}
      \label{tab:rgb-l1-crello}
  \small
  \centering
  \begin{tabular}{@{}ll r cccc c@{}}
    Layers & Variant & $m$
      & Mean $\downarrow$ & Pred\,0 $\downarrow$ & Pred\,1 $\downarrow$ & Pred\,2 $\downarrow$ &  Pred\,3 $\downarrow$ \\
    \midrule
    \multirow{2}{*}{$L{=}2$}
      & Qwen-Image-Layered         &  3 & 0.1706          & 0.2938 & \best{0.0473} & ---    & ---    \\
      & \method{} &  3 & \best{0.1635} & \best{0.2511} & 0.0760 & ---    & ---    \\
    \midrule
    \multirow{2}{*}{$L{=}3$}
      & Qwen-Image-Layered         & 29 & 0.0879          & 0.0734 & 0.0938 & \best{0.0965} & ---    \\
      & \method{} & 29 & \best{0.0767} & \best{0.0502} & \best{0.0786} & 0.1012 & ---    \\
    \midrule
    \multirow{2}{*}{$L{=}4$}
      & Qwen-Image-Layered         & 90 & 0.0712          & 0.0954 & 0.0848 & 0.0590 & \best{0.0457} \\
      & \method{} & 90 & \best{0.0660} & \best{0.0795} & \best{0.0678} & \best{0.0573} & 0.0594 \\
  \end{tabular}
\end{table}

\vspace{-0.5em}

\paragraph{Comparison with LayerD.}
LayerD~\citep{suzuki2025layerd} and \method{} take different
approaches to decomposition uncertainty. LayerD is conservative:
when separation is hard, it tends to return the input largely
intact as a single layer, producing fewer but more confident
outputs. \method{} always populates the requested number of
layers. The two strategies have different downstream implications,
and \Cref{tab:layerd-comparison} reflects this: \method{} achieves
substantially higher distribution evenness
because it actually fills the requested slots with distinct
content, while LayerD scores marginally higher on Layer~0 quality
because an unmodified copy of the input is
by construction a plausible scene. For most editing workflows---
where the value of a decomposition is in having usable separate
layers---\method{}'s behaviour is the more useful one; the Layer~0
gap reflects an artifact of the metric rewarding faithful pixel
content rather than a genuine quality advantage.

\begin{table}[t]
  \caption{Comparison on the held-out LAION-Aesthetics set
    ($n{=}480$) at four output layers. LayerD frequently returns
    fewer than four layers; unfilled slots are padded with empty
    layers for metric computation. Higher is better for both
    columns; bold marks the best per column.}
  \label{tab:layerd-comparison}
  \centering
  \small
  \begin{tabular}{@{}lcc@{}}
    Variant & Distrib.\ $\uparrow$ & Layer~0 Q $\uparrow$ \\
    \midrule
    Qwen-Image-Layered & 0.5282 & 0.3817 \\
    \method{}          & \best{0.7339} & 0.6148 \\
    LayerD             & 0.0585 & \best{0.7136} \\
  \end{tabular}
\end{table}

\subsection{Ablation: Effect of Text Conditioning}
\label{sec:ablation-textcond}

We compare two fixed prompts applied uniformly across all training images: a \emph{basic} prompt and a \emph{detailed} prompt mirroring the reward rubric's evaluation axes (full text in \Cref{app:ablation-prompts}). Ablation runs use $G{=}8$, all other hyperparameters match \Cref{app:training-details}.

The detailed-prompt run underperforms the main run on every axis. Bad
layers fall more slowly, feature distribution evenness plateaus lower, and Layer~0 quality actively degrades from
${\sim}0.44$ to ${\sim}0.32$ where the main run improves from
${\sim}0.40$ to ${\sim}0.74$. Conditioning on a prompt that describes
an idealised multi-object scene may give the policy model a sense of direction the VLM judge might evaluate.

\begin{table}[t]
\centering
\small
\setlength{\tabcolsep}{4pt}
\caption{Text conditioning ablation across training steps (480-image
  eval). Comparing a basic fixed prompt against a detailed fixed
  prompt mirroring the reward rubric, both applied uniformly across
  training images. ``Bad'' is the average number of blank or
  over-glazed layers per generation; \emph{Distrib.} is feature
  distribution evenness; \emph{L0~Q.} is Layer~0 quality. Arrows
  indicate desired direction.}
\label{tab:textcond_ablation}
\begin{tabular}{r|ccc|ccc}
 & \multicolumn{3}{c|}{Basic Prompt} & \multicolumn{3}{c}{Detailed prompt} \\
\cmidrule(lr){2-4} \cmidrule(lr){5-7}
Step & Bad$\downarrow$ & Distrib.$\uparrow$ & L0\,Q.$\uparrow$
     & Bad$\downarrow$ & Distrib.$\uparrow$ & L0\,Q.$\uparrow$ \\
\midrule
0   & 1.600 & 0.526 & 0.403 & 1.692 & 0.536 & 0.435 \\
40  & 0.792 & 0.631 & 0.517 & 1.433 & 0.660 & 0.349 \\
80  & 0.368 & 0.692 & 0.554 & 1.267 & 0.713 & 0.369 \\
120 & 0.245 & 0.726 & 0.573 & 1.362 & 0.688 & 0.316 \\
160 & 0.226 & 0.739 & 0.643 & 1.015 & 0.690 & 0.297 \\
200 & 0.335 & 0.733 & 0.738 & 0.612 & 0.691 & 0.323 \\
\end{tabular}
\vspace{-1em}
\end{table}

\begin{table}[t]
\centering
\small
\setlength{\tabcolsep}{4pt}
\caption{Calibration ablation across training steps (480-image eval).
  ``Bad'' is the average number of blank or over-glazed layers per
  generation; \emph{Qual.}, \emph{Sharp.}, and \emph{SSIM} are
  Layer~0 quality metrics.}
\label{tab:grid_ablation_longer}
\begin{tabular}{r|cccc|cccc}
 & \multicolumn{4}{c|}{No Calibration} & \multicolumn{4}{c}{With Calibration} \\
\cmidrule(lr){2-5} \cmidrule(lr){6-9}
Step & Qual.$\uparrow$ & Sharp.$\uparrow$ & SSIM$\uparrow$ & Bad$\downarrow$
     & Qual.$\uparrow$ & Sharp.$\uparrow$ & SSIM$\uparrow$ & Bad$\downarrow$ \\
\midrule
0   & 0.611 & 0.236 & 0.579 & 1.008 & 0.611 & 0.236 & 0.579 & 1.008 \\
40  & 0.585 & 0.189 & 0.500 & 0.600 & 0.595 & 0.198 & 0.495 & 0.581 \\
80  & 0.560 & 0.154 & 0.437 & 0.346 & 0.590 & 0.184 & 0.489 & 0.560 \\
120 & 0.564 & 0.139 & 0.420 & 0.298 & 0.598 & 0.198 & 0.514 & 0.500 \\
160 & 0.577 & 0.164 & 0.444 & 0.392 & 0.618 & 0.238 & 0.556 & 0.627 \\
200 & 0.587 & 0.205 & 0.522 & 0.658 & 0.637 & 0.278 & 0.548 & 0.398 \\
\end{tabular}
\vspace{-1em}
\end{table}
\vspace{-1em}

\subsection{Ablation: Effect of Grid Calibration}
\label{sec:ablation-calibration}

To isolate the contribution of the relative grid calibration phase
(\Cref{sec:method-reward-calibration}), we train two otherwise
identical runs---one with the full two-phase reward, one with
Phase~1 individual scoring alone---and evaluate every 40 steps on a
held-out set of 480 LAION-Aesthetics
images~\citep{schuhmann2022laion5b}. Results are in
\Cref{tab:grid_ablation_longer}.
\vspace{-0.5em}
\paragraph{Bad layer reduction is largely unaffected.}
Both runs reduce bad layers from $1.008$ to $0.4$--$0.7$ across
mid-training, with neither variant consistently leading. Phase~1's
content-validity and alpha-cleanliness criteria already produce
enough variance on these binary-like defects for GRPO to learn from.
\vspace{-0.5em}
\paragraph{Image quality benefits from calibration.}
All three Layer~0 quality metrics separate the two runs from step~80
onward. SSIM averages $0.52$ (calibrated) vs.\ $0.45$ (uncalibrated)
across steps 80--200; combined quality and edge sharpness show the
same pattern. The gap matches the score-compression hypothesis
behind Phase~2: when all candidates in a group are non-degenerate,
the remaining quality differences (inpainting plausibility, edge
quality) compress into a narrow Phase~1 band, yielding near-uniform
advantages. Forcing explicit relative judgments restores the
within-group variance the policy update requires. More broadly,
coarse failure modes are well-captured by absolute scoring;
fine-grained perceptual quality benefits from relative calibration.

\vspace{-1em}

\section{Conclusion}
\label{sec:conclusion}
\vspace{-0.5em}
We have presented \method{}, a method for improving image layer
decomposition models through reinforcement learning with
VLM-provided rewards. Combining Flow-GRPO's SDE-augmented policy
optimisation with a two-phase VLM scoring protocol---structured
per-sample evaluation followed by relative grid calibration---we
convert black-box judge feedback into a learning signal
discriminative enough to drive fine-grained improvements in layer
separation, content validity, and feature distribution, without
task-specific reward training, synthetic targets, or human
annotation.
\vspace{-0.5em}

The recipe generalises: any conditional generator whose outputs can
be meaningfully evaluated by a VLM (style transfer, inpainting,
relighting, scene rearrangement) could in principle be fine-tuned
with the loop. Promising extensions
include replacing the grid calibration's free-form numeric output
with logit-based pairwise preferences (avoiding the failure mode
TOPReward~\citep{topreward2026} targets) and automating rubric
design itself by having a VLM critique its own scoring criteria.
\vspace{-0.5em}
\paragraph{Limitations.}
\method{} relies on a proprietary VLM as the reward model, which
introduces API cost per training step and a dependency on a specific
model snapshot whose score distribution may drift across versions. Our evaluation
relies on automated metrics and qualitative inspection rather than
human studies; the metrics correlate with editing usefulness but do
not directly measure it. Finally, training was capped at five layers
per sample for compute reasons, so behaviour on high-layer-count
decompositions (the base model supports up to 20) is not directly
evaluated.


\bibliographystyle{ieeenat_fullname}
\bibliography{references}

\clearpage
\appendix
\crefalias{section}{appendix}

\section{Algorithm Pseudocode}
\label{app:algorithm}

\begin{algorithm}[ht]
  \caption{\method{} Training Step (adapted from Flow-GRPO with GRPO-Guard stabilisation)}
  \label{alg:flow-grpo}
  \begin{algorithmic}[1]
    \Require Policy $\pi_\theta$ (LoRA), reference $\pi_{\mathrm{ref}}$,
      reward model $R$, group size $G$, latent dimensionality $D$,
      advantage clip $c_{\mathrm{adv}}$, ratio clip $\epsilon_c$,
      KL weight $\beta$, learning rate $\eta$
    \State Sample condition image $x_{\mathrm{cond}}$ and $x_{\mathrm{text}}$ prompt.
    \State Encode condition: $z_{\mathrm{cond}} \gets \mathrm{VAE.encode}(x_{\mathrm{cond}})$+ $x_{\mathrm{text}}$.
    \Statex \textbf{// Phase 1: Generate}
    \For{$g = 1, \ldots, G$}
      \State $z_T^{(g)} \sim \mathcal{N}(0, I)$
      \State $(z_0^{(g)},\; \tau^{(g)},\; \log\pi_{\theta_\mathrm{old}}^{(g)})
        \gets \mathrm{SDE\_Generate}(z_T^{(g)}, z_{\mathrm{cond}})$
      \State $\mathrm{layers}^{(g)} \gets \mathrm{VAE.decode}(z_0^{(g)})$
    \EndFor
    \Statex \textbf{// Phase 2: Score}
    \For{$g = 1, \ldots, G$}
      \State $r^{(g)} \gets R\bigl(\operatorname{Composite}(\mathrm{layers}^{(g)}),\, \mathrm{layers}^{(g)}\bigr)$
    \EndFor
    \State $\hat{A}^{(g)} \gets \mathrm{clip}\!\left(\frac{r^{(g)} - \bar{r}}{\sigma_r + \nu},\; -c_{\mathrm{adv}},\; c_{\mathrm{adv}}\right)$
    \Statex \textbf{// Phase 3: Train}
    \For{$g = 1, \ldots, G$}
      \For{each SDE step $i$ in $\tau^{(g)}$}
        \State $\log\pi_\theta^{(g,i)},\; \log\pi_{\mathrm{ref}}^{(g,i)} \gets
          \mathrm{Replay}(\tau^{(g)}_i, \pi_\theta, \pi_{\mathrm{ref}})$
        \State $\log\rho^{(g,i)} \gets
          \Bigl(\!\textstyle\sum_d \log\pi_\theta^{(g,i)}[d]
              - \sum_d \log\pi_{\theta_\mathrm{old}}^{(g,i)}[d]\Bigr)
          \;/\; \sqrt{D}$
          \Comment{sum-and-rescale RatioNorm}
        \State $\log\rho^{(g,i)} \gets \bigl(\log\rho^{(g,i)} - \mu_i\bigr) / s_i$
          \Comment{per-step centring (mean $\mu_i$, scale $s_i$ across group)}
        \State $\rho^{(g,i)} \gets \exp\!\bigl(\log\rho^{(g,i)}\bigr)$
        \State $\widehat{\KL}^{(g,i)} \gets
          \log\pi_\theta^{(g,i)} - \log\pi_{\mathrm{ref}}^{(g,i)}$
        \State $\mathcal{L}^{(g,i)} \gets
          -\frac{1}{\Delta t_i}\min\!\bigl(
            \rho^{(g,i)} \hat{A}^{(g)},\;
            \mathrm{clip}(\rho^{(g,i)}, 1{-}\epsilon_c, 1{+}\epsilon_c)\,
            \hat{A}^{(g)}
          \bigr)
          + \beta\, \widehat{\KL}^{(g,i)}$
          \Comment{Gradient reweight by $1/\Delta t$}
      \EndFor
    \EndFor
    \State $\theta \gets \theta - \eta \nabla_\theta
      \frac{1}{G}\sum_g \frac{1}{|\tau^{(g)}|}\sum_i \mathcal{L}^{(g,i)}$
  \end{algorithmic}
\end{algorithm}

\section{Reward Prompt Template}
\label{app:reward-prompt}

This appendix provides the exact prompts sent to the VLM reward model
(\texttt{gemini-3-flash-preview}) during training.

\paragraph{Reward model version.}
All reward model calls use \texttt{gemini-3-flash-preview} via the
Google AI Studio API, pinned to the model snapshot available between October 2025 and the
date of submission. Researchers
extending the method should expect score distributions to drift
across model versions; we recommend re-calibrating the Phase~1
anchor descriptions when switching reward models.

\subsection*{B.1\quad System Prompt}

The following system message is prepended to all reward model calls:

\begin{quote}
\small
\texttt{You are an expert image compositor evaluating layer decomposition
quality. You will see an original composite image and its decomposition into
separate layers. Layer 0 is ALWAYS the background flat. The remaining layers
are foreground elements.}

\texttt{Score ONLY based on what you see. Be harsh and discriminating -- give
different scores to samples that genuinely differ in quality. Do NOT give
identical scores to all samples. Respond ONLY with valid JSON, no other text.}
\end{quote}

\subsection*{B.2\quad Phase 1: Individual Scoring Prompt}

Each sample is presented as the composite image followed by $N$ layer images
(composited onto white backgrounds), with the following instruction appended.
The placeholder \texttt{\{num\_layers\}} is replaced with the number of layers
for that training step.

\begin{quote}
\small
\sloppy

\texttt{Score this \{num\_layers\}-layer decomposition. Layer 0 is the
background; layers 1+ are foreground.}

\texttt{CRITERIA (score each 0-5):}

\texttt{1. semantic\_separation (0-5): Each foreground layer should contain ONE
distinct, complete object or semantic element (e.g.\ a person, a car, a tree).
Score 0 if a single object is arbitrarily split across multiple layers or if
layers contain random crops/slices of the scene rather than meaningful
elements. Score 5 if every foreground layer isolates a complete, distinct
object and no object is split across layers.}

\texttt{2. alpha\_cleanliness (0-5): Foreground layers should have crisp,
binary-like alpha with clean edges. Score 0 if layers show a semi-transparent
haze, ghosting, colour bleed, or a milky/glazed wash over areas that should be
fully transparent. Transparent regions must be FULLY transparent with zero
colour residue. Score 5 if alpha masks are sharp, edges are clean, and
transparent regions are completely clear with no residual colour or haze.}

\texttt{3. background\_inpainting (0-5): Layer 0 (the background) should look
like a plausible complete scene with foreground objects removed and their
regions filled in convincingly. Score 0 if the background is blurry, has
obvious holes, smeared patches, or copy-paste artifacts where foreground
objects were removed. Score 5 if the inpainted regions blend seamlessly with
the surrounding background, maintaining consistent texture, lighting, and
detail.}

\texttt{4. feature\_distribution (0-5): Visual content should be meaningfully
spread across layers. Score 0 if most content is crammed into one layer while
others are blank or near-empty. Score 5 if layers have a balanced, meaningful
distribution of the scene's content.}

\texttt{5. content\_validity (0-5): Penalize blank, empty, or noise-only
layers. Score 0 if most layers are blank or contain only noise/blur. Score 5 if
all layers have clear, recognizable content.}

\texttt{- total (0-25): Sum of all five scores.}

\texttt{Return ONLY valid JSON:
\{"semantic\_separation":X, "alpha\_cleanliness":Y, "background\_inpainting":Z, "feature\_distribution":W, "content\_validity":V, "total":T\}}
\end{quote}

\subsection*{B.3\quad Phase 2: Grid Calibration Prompt}

\paragraph{Grid construction.}
The $G$ Phase~1 RGB composites are arranged left-to-right, top-to-bottom
in a $\lceil\sqrt{G}\rceil \times \lceil G/\lceil\sqrt{G}\rceil\rceil$
tiling on a white canvas (for our default $G{=}16$, a $3{\times}2$ grid).
Each cell is rendered at $320{\times}320$ with a uniform white margin
between cells, and the integer index $0,\ldots,G{-}1$ is rasterised in
the top-left corner of each cell as a black sans-serif numeral on a
white background. The completed grid is sent to the VLM as a single
PNG.

\paragraph{Prompt.}
The grid is sent alongside the following instruction. Placeholders
\texttt{\{G\}}, \texttt{\{Gm1\}}, and \texttt{\{scores\_csv\}} are
replaced with the group size, $G{-}1$, and the comma-separated Phase~1
scores respectively.

\begin{quote}
\small
\texttt{The grid shows \{G\} layer-decomposition samples arranged
left-to-right, top-to-bottom, labeled 0-\{Gm1\}.}

\texttt{Initial individual scores: \{scores\_csv\}}

\texttt{Re-score each sample RELATIVE to the others. Give higher scores to
better decompositions and lower to worse ones. Be discriminating -- spread the
scores. Pay special attention to:}

\texttt{- Which samples keep whole objects on single layers vs.\ splitting them?}

\texttt{- Which samples have that semi-transparent glaze/ghosting vs.\ clean alpha?}

\texttt{- Which samples have convincing background inpainting vs.\ blurry fills?}

\texttt{Reply with ONLY \{G\} comma-separated decimal values in [0,1], one per
sample in order:}
\end{quote}

\section{Additional Qualitative Examples}
\label{app:qualitative}

\begin{figure}[p]
  \centering
  \includegraphics[width=\linewidth,height=0.8\textheight,keepaspectratio]{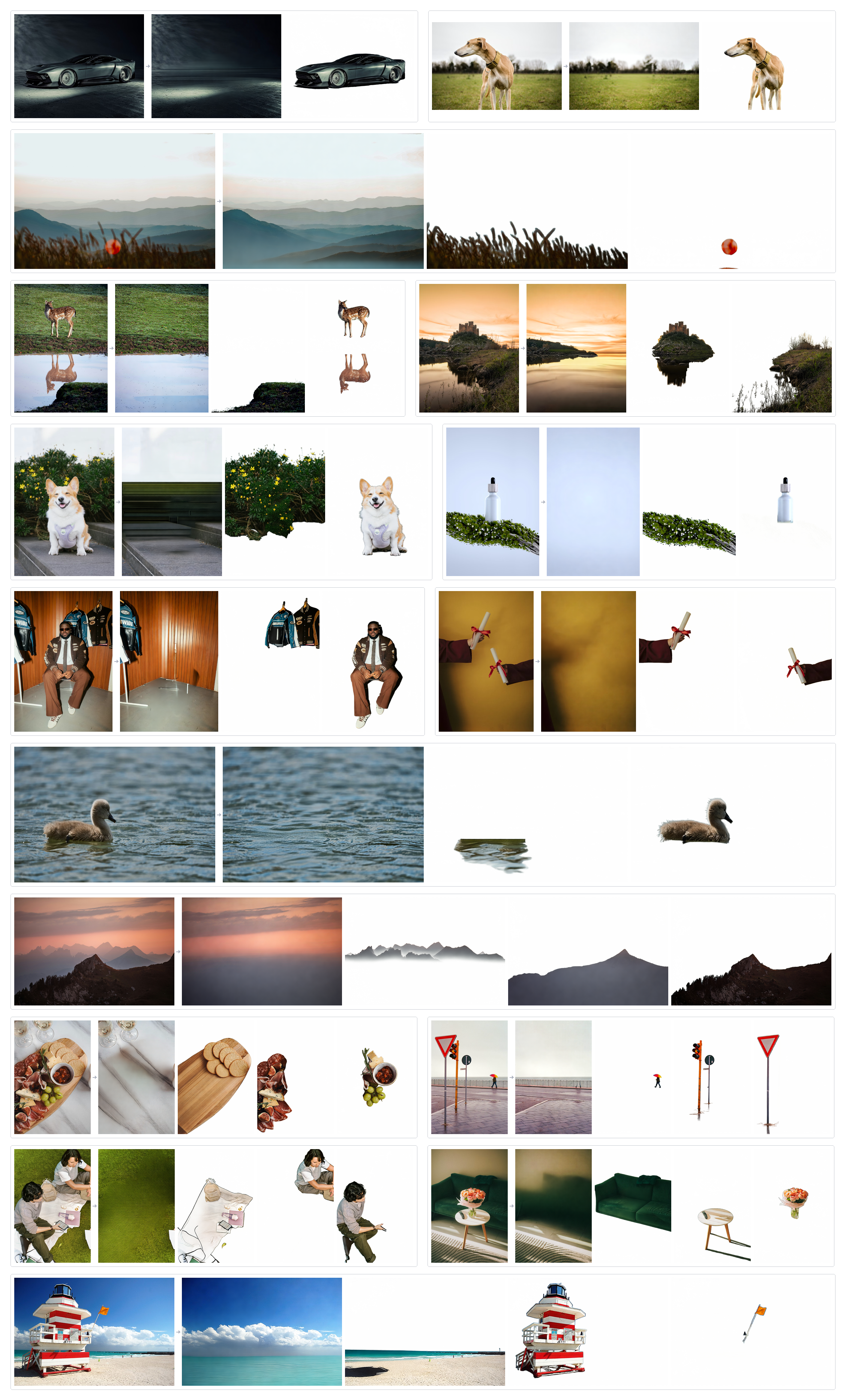}
  \titlecaption{Extended qualitative gallery}{%
    Layer decompositions from the \method{}-fine-tuned model on a
    diverse set of held-out inputs. Each row shows the reconstructed composite and the individual layers composited
    onto white backgrounds. Examples illustrate consistent behaviour
    across subject matter: clean isolation of foreground objects
    (corgi, swan, deer, jacket), plausible background inpainting
    where foreground elements are removed (lighthouse, statue island,
    mountain scene), and reasonable separation of multiple
    foreground elements onto distinct layers (charcuterie board,
    yield-sign scene, couch and side table).}
  \label{fig:qualitative-gallery}
\end{figure}

\Cref{fig:qualitative-gallery} presents an extended gallery of layer
decompositions produced by the \method{}-fine-tuned model on held-out
images from LAION-Aesthetics, spanning a range of subject matter:
natural photographs (landscapes, wildlife, portraits), studio product
shots, automotive renders, and scene compositions with varying
foreground complexity.

\section{Additional Calibration Ablation Metrics}
\label{app:calibration-extra}

\begin{figure}[h]
    \centering
    \includegraphics[width=1.0\linewidth]{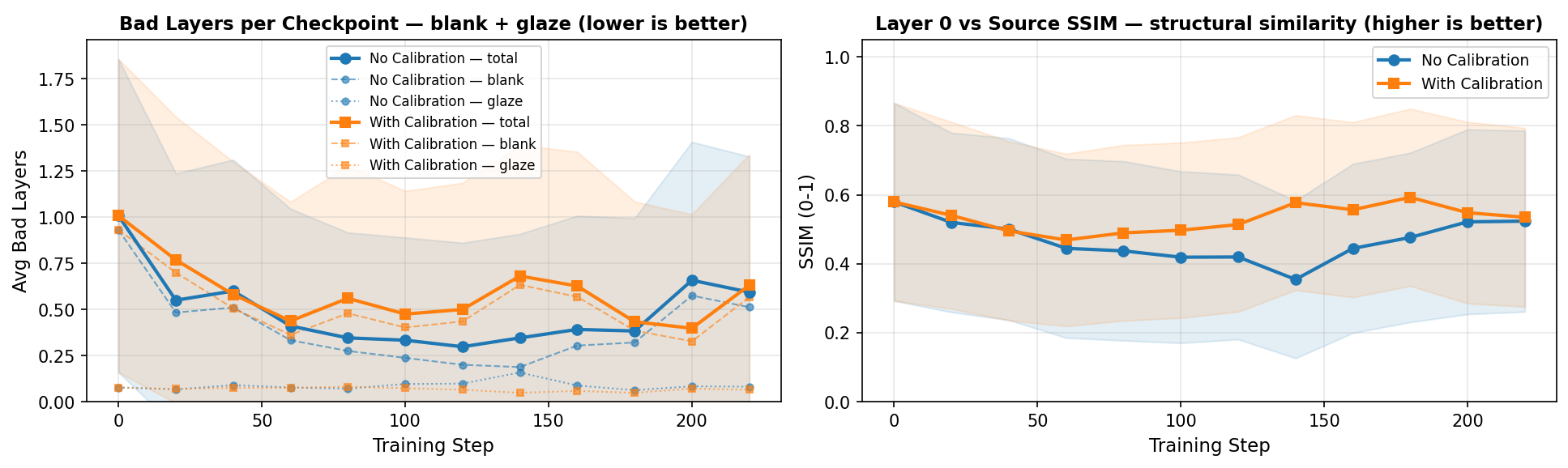}
    \titlecaption{Calibration ablation (additional Layer~0 metrics)}{%
      Layer~0 combined quality (left) and edge-density sharpness
      (right) over training, comparing the full two-phase reward
      with grid calibration against Phase~1 individual scoring alone.
      Both metrics show the calibrated run maintaining a small but
      consistent lead from approximately step~120 onward,
      corroborating the SSIM result reported in
      \Cref{tab:grid_ablation_longer}.}
    \label{fig:ablation-calibration-extra}
\end{figure}

\Cref{fig:ablation-calibration-extra} reports two additional Layer~0
quality metrics for the calibration ablation of
\Cref{sec:ablation-calibration}: a combined quality score and an
edge-density sharpness measure. Both show the same qualitative
pattern as the SSIM result in the main text---the calibrated run
maintains a small but consistent lead over the uncalibrated run from
the mid-training checkpoints onward---providing converging evidence
that grid calibration improves fine-grained background quality
without affecting bad-layer reduction.

\section{Architecture Details}
\label{app:architecture}

The base \modelname{}~\citep{yin2025qwen} is a flow-matching
transformer~\citep{liu2022flowa, liu2022flowb} that produces $N$-layer
RGBA decompositions conditioned on an input image. The architecture
comprises:
\begin{itemize}[leftmargin=*]
  \item A 3D variational autoencoder (VAE) that encodes 4-channel
    RGBA frames with $8{\times}$ spatial compression into a 16-channel
    latent space.
  \item A sequence-based transformer operating on $2{\times}2$
    patch-packed latents, yielding token dimension
    $16 \times 4 = 64$.
  \item A text encoder for prompt conditioning.
\end{itemize}
The condition image is encoded through the same VAE pipeline and
concatenated along the sequence dimension of the transformer input,
with per-frame spatial metadata provided to the attention mechanism
to distinguish generated layer tokens from conditioning tokens.

We apply Low-Rank Adaptation~\citep{hu2021lora} with rank $r{=}16$
and $\alpha{=}16$ to all attention projection layers and feed-forward
layers, keeping all other parameters frozen.

\section{Training Implementation Details}
\label{app:training-details}

\paragraph{SDE schedule and CFG.}
During the generation phase we use a reduced schedule of
$T_{\mathrm{train}}$ SDE steps (typically $8$, compared to $50$ at
inference) to keep memory and compute costs tractable across $G$
group samples. Following Flow-GRPO, classifier-free guidance is
disabled during training ($\text{CFG} = 1.0$, halving the number of
forward passes per step) but enabled at evaluation
($\text{CFG} = 4.0$); \citet{flowgrpo2025} found that this asymmetry
did not degrade final sample quality while substantially reducing
training cost.

\paragraph{Trajectory replay.}
For each stored SDE step $i$, the current policy's transition mean
$\mu_i^\theta$ is recomputed via a forward pass through the
LoRA-adapted transformer with gradients enabled. The KL reference
is computed by running the same forward pass with LoRA adapters
disabled, yielding the pre-adaptation base model's prediction at
zero additional memory cost.

\paragraph{Ratio normalisation and gradient reweighting.}
We adopt the GRPO-Guard~\citep{grpoguard2025} stabilisation scheme
(\Cref{sec:bg-grpoguard}) with one modification. GRPO-Guard's
default RatioNorm computes a spatial mean of per-element
log-probabilities scaled by the noise standard deviation. For
\modelname{}, this suppresses the magnitudes of the log-ratio to
near zero: the model packs multiple RGBA layers into a single
latent sequence at $640 \times 640$ resolution, producing a
dimensionality per-step~$D$ considerably higher than the
single-image $512 \times 512$ latents of SD3.5 on which GRPO-Guard
was developed. We therefore compute the log-probability per-step as
a sum over spatial dimensions and normalise by $\sqrt{D}$ before
applying GRPO-Guard's per-step centring (\Cref{alg:flow-grpo},
lines 15--16), preserving $\mathcal{O}(1)$ ratio magnitudes while
retaining RatioNorm's centring and variance-stabilisation
properties. The per-step policy loss is additionally scaled by
$\delta = 1/\Delta t$ (gradient reweighting) to equalize gradient
magnitudes across the noise schedule, following GRPO-Guard without
modification.

\paragraph{Training and compute.}
$K{=}1$ PPO-style epoch per round, $G{=}16$, $\epsilon_c{=}0.2$, $\beta{=}10^{-3}$, AdamW with $\eta{=}10^{-5}$, advantage clip $c_{\mathrm{adv}}{=}5.0$, gradient clip $\|\nabla\|_{\max}{=}1.0$. The main $600$-step run was trained on $8{\times}$ NVIDIA H200 GPUs in ${\sim}48$ hours.

\section{Data Preprocessing}
\label{app:data}

The primary source is Fine-T2I~\citep{ma2026finet2i}, an
aesthetically filtered subset of photographs and artworks. All
images are resized to $640 \times 640$ and normalised to $[-1, 1]$;
associated captions serve as text prompts for the model's text
conditioning. Images are shuffled at each epoch. The number of
output layers per training step is sampled uniformly from
$[\text{min\_layers}, \text{max\_layers}]$ (typically $[2, 5]$),
exposing the model to variable decomposition complexity.

\section{LayerD Comparison Setup}
\label{app:layerd-setup}

We compare against LayerD~\citep{suzuki2025layerd} on the held-out
LAION-Aesthetics set used in \Cref{fig:eval-metrics}. LayerD's
output count is variable and frequently smaller than the four
layers our metrics score---the method tends to leave the input
largely intact rather than separating it, sometimes returning a
single layer containing the full image. To make the metrics
computable on a fixed slot count, we pad LayerD's outputs with
empty (white) layers up to four. We treat empty slots literally:
for downstream editing purposes, an unfilled slot is an empty
slot, and the resulting scores characterise the difference in
decomposition strategy rather than quality on a shared axis (see
\Cref{sec:results-quantitative}).

\begin{figure}[t]
  \centering
  \includegraphics[width=0.85\linewidth]{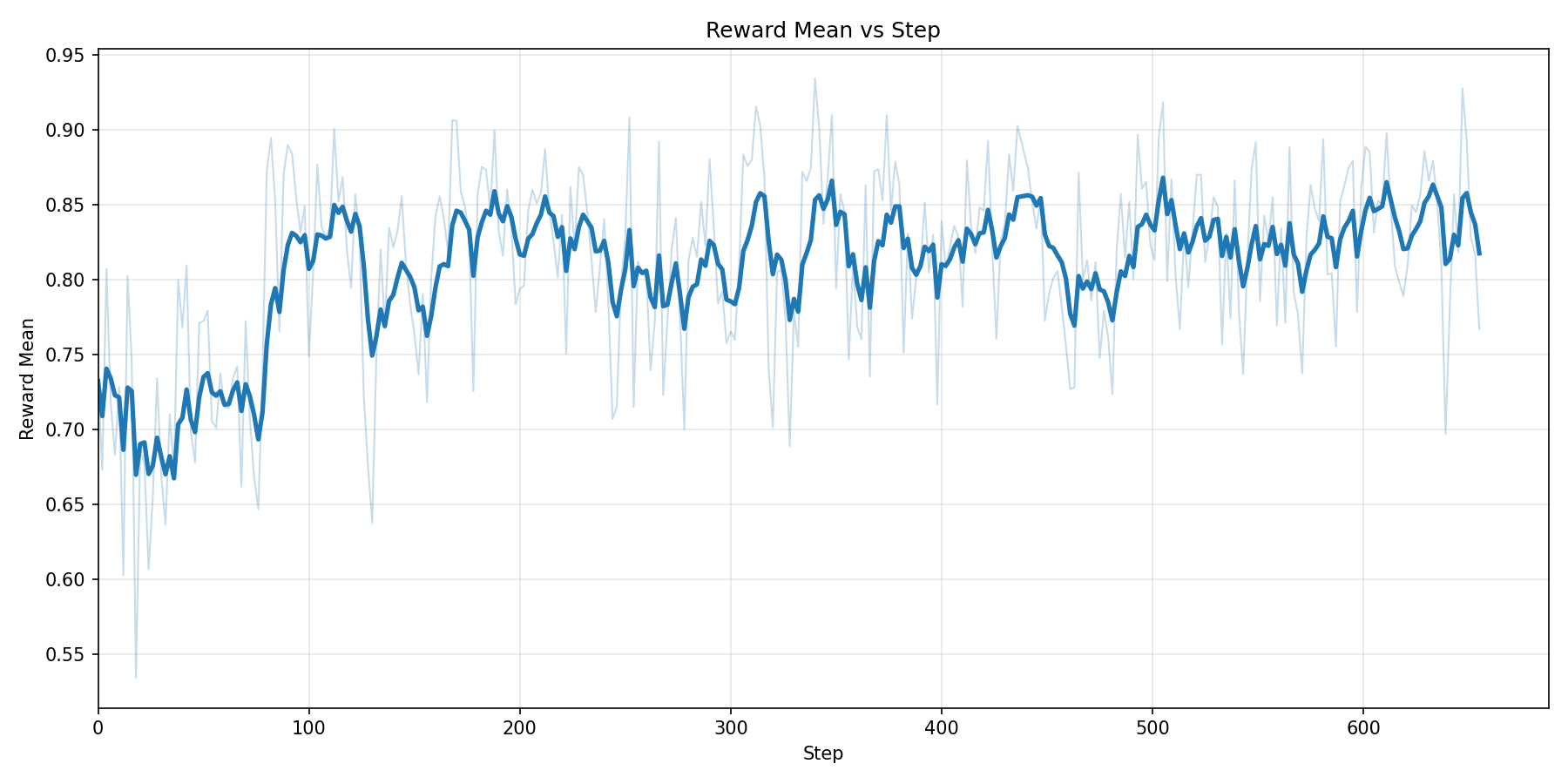}
    \titlecaption{Mean VLM reward during training}{
    Phase~2 calibrated reward per step (light) and rolling
    average (dark).  The mean reward rises over the first ${\sim}100$ steps as the policy
    eliminates the worst failure modes, then plateaus around
    $0.83$--$0.85$ with high per-step variance.  Under GRPO's
    within-group normalisation (\Cref{eq:grpo-advantage}), learning
    requires only relative discrimination among group members, not
    rising absolute scores; the residual variance reflects
    conditioning-image difficulty rather than policy quality.}
  \label{fig:reward-progression}
\end{figure}

\section{Crello Evaluation Metric}
\label{app:crello-metric}

\citet{yin2025qwen} measure per-layer reconstruction quality on the
Crello dataset~\citep{yamaguchi2021canvasvae} by comparing each
predicted layer against the corresponding ground-truth layer at the
same index, using the LayerD~\citep{suzuki2025layerd} evaluation
protocol with order-aware Dynamic Time Warping. Since our RL reward
signal may encourage the model to reorder layers relative to the base
model's conventions (e.g., placing the most prominent foreground
element on layer~1 rather than layer~2), a fixed-index comparison
would penalise semantically correct decompositions that simply assign
layers in a different order. We therefore modify the metric to use
\emph{best-match} assignment: for each reference layer, we select the
predicted layer with the highest RGB similarity and compute the
reconstruction error against that match, rather than relying on
positional correspondence. This isolates decomposition quality from
layer ordering, ensuring that improvements in semantic separation and
content distribution are not masked by index-level misalignment
introduced by the reward signal.

\paragraph{Held-out evaluation set.}
We evaluate on a fixed set of 480 images sampled from
LAION-Aesthetics~\citep{schuhmann2022laion5b} and held constant
across all checkpoints and ablations. The evaluation set is
fully disjoint from the training data, which is drawn from a
separate  dataset (described in \Cref{sec:method-data});
no LAION-Aesthetics images appear in training. 

\section{Text Conditioning Ablation Prompts}
\label{app:ablation-prompts}

The text conditioning ablation in \Cref{sec:ablation-textcond}
compares two fixed prompts applied uniformly across all training
images, replacing the per-image dataset captions used in the main
run. The exact prompt strings are:

\begin{description}[leftmargin=1.5em,style=nextline]
  \item[Basic prompt.]
    \texttt{a clean, well composed image.}
  \item[Detailed prompt.]
    \texttt{a high quality image with multiple distinct objects
    clearly separated from a clean background, sharp edges, vivid
    colors, balanced lighting, well-defined foreground elements
    against a coherent backdrop, professional composition with
    clear depth layers.}
\end{description}

The detailed prompt was chosen to mirror the reward rubric's
evaluation axes (object separation, alpha cleanliness, background
coherence, feature distribution), testing whether prompt-rubric
alignment helps or hinders training. As reported in
\Cref{sec:ablation-textcond}, the detailed prompt fails to reduce
bad layers despite this surface alignment.


\ifshowchecklist
\clearpage
\section*{NeurIPS Paper Checklist}

\begin{enumerate}

\item {\bf Claims}
    \item[] Question: Do the main claims made in the abstract and introduction accurately reflect the paper's contributions and scope?
    \item[] Answer: \answerYes{} 
    \item[] Justification: We clearly stated that investigated RL fine-tuning an editing model to improve the performance and demonstrated it with thorough ablations (\Cref{sec:ablation-textcond,sec:ablation-calibration}) and comparisons (\Cref{sec:results-qualitative}).

\item {\bf Limitations}
    \item[] Question: Does the paper discuss the limitations of the work performed by the authors?
    \item[] Answer: \answerYes{} 
    \item[] Justification: They are clearly stated in \Cref{sec:conclusion}.

\item {\bf Theory assumptions and proofs}
    \item[] Question: For each theoretical result, does the paper provide the full set of assumptions and a complete (and correct) proof?
    \item[] Answer: \answerYes{} 
    \item[] Justification: We apply existing concepts in novel ways and for novel tasks.

    \item {\bf Experimental result reproducibility}
    \item[] Question: Does the paper fully disclose all the information needed to reproduce the main experimental results of the paper to the extent that it affects the main claims and/or conclusions of the paper (regardless of whether the code and data are provided or not)?
    \item[] Answer: \answerYes{} 
    \item[] \item[] Justification: The method, architecture, hyperparameters, and
      reward prompts are described in sufficient detail for reproduction
      in \Cref{sec:method,app:training-details,app:reward-prompt}, and the
      base model and dataset are publicly available. We plan to release the trained LoRA adapters.

\item {\bf Open access to data and code}
    \item[] Question: Does the paper provide open access to the data and code, with sufficient instructions to faithfully reproduce the main experimental results, as described in supplemental material?
    \item[] Answer: \answerYes{} 
    \item[] Justification: Code release is under consideration. The
      reward prompts (\Cref{app:reward-prompt}), training procedure
      (\Cref{app:training-details}), and dataset and base model
      (\Cref{sec:method-data,sec:method-architecture}) are documented in
      full to enable reproduction without code release.

\item {\bf Experimental setting/details}
    \item[] Question: Does the paper specify all the training and test details (e.g., data splits, hyperparameters, how they were chosen, type of optimizer, etc.) necessary to understand the results?
    \item[] Answer: \answerYes{} 
    \item[] Justification: We provided detailed documentation of our setup in the method section \Cref{sec:method}.

\item {\bf Experiment statistical significance}
    \item[] Question: Does the paper report error bars suitably and correctly defined or other appropriate information about the statistical significance of the experiments?
    \item[] Answer: \answerYes{} 
    \item[] Justification:   All evaluation plots include shaded error bands representing $\pm 1$ standard deviation ($1\sigma$)
  computed across all test images within each checkpoint variant. For example, a checkpoint evaluated on 480 images
  reports the mean metric value with the standard deviation of that metric across the 480 per-image scores. The error
  bands capture variability due to input image diversity (different SVG complexities, content types, and colour
  distributions) under fixed model weights and sampling parameters. Standard deviations are computed directly. We do not assume normally distributed errors; the
  bands are shown as symmetric $\pm 1\sigma$ for visual clarity, but we note that metrics bounded in $[0,1]$ may have
  asymmetric tails near the boundaries.

\item {\bf Experiments compute resources}
    \item[] Question: For each experiment, does the paper provide sufficient information on the computer resources (type of compute workers, memory, time of execution) needed to reproduce the experiments?
    \item[] Answer: \answerYes{}{} 
    \item[] Justification: The main run used 8$\times$ NVIDIA H200 GPUs
      for approximately 48 hours (600 steps). Ablation runs used the
      same hardware with fewer steps. Total compute including
      preliminary experiments exceeded the reported runs by
      approximately 3$\times$. Details are provided in
      \Cref{sec:experiments}.
    
\item {\bf Code of ethics}
    \item[] Question: Does the research conducted in the paper conform, in every respect, with the NeurIPS Code of Ethics \url{https://neurips.cc/public/EthicsGuidelines}?
    \item[] Answer: \answerYes{} 
    \item[] Justification: We conform with the Code of Ethics fully.

\item {\bf Broader impacts}
    \item[] Question: Does the paper discuss both potential positive societal impacts and negative societal impacts of the work performed?
    \item[] Answer: \answerYes{}
    \item[] Justification: \method{} improves the quality of automated layer
      decomposition, which lowers the technical barrier for image
      compositing and editing. Beneficial uses include accessibility tooling,
      education, and creative workflows. The same capability could marginally
      ease the production of misleading composite imagery, though the model
      operates on existing images rather than synthesising them and does not
      provide capabilities beyond those of existing editing tools. The reward
      signal incorporates a penalty on unsafe content during training, which
      is expected to reduce the prevalence of such outputs relative to the
      base model.
    
\item {\bf Safeguards}
    \item[] Question: Does the paper describe safeguards that have been put in place for responsible release of data or models that have a high risk for misuse (e.g., pretrained language models, image generators, or scraped datasets)?
    \item[] Answer: \answerYes{}
    \item[] Justification: The model performs decomposition of
      user-supplied images rather than open-ended synthesis, which
      limits the misuse surface relative to general-purpose image
      generators. The reward signal additionally penalises unsafe
      content during training. Release plans for the trained adapters
      have not yet been finalised; any release would be accompanied by
      appropriate safeguards.

\item {\bf Licenses for existing assets}
    \item[] Question: Are the creators or original owners of assets (e.g., code, data, models), used in the paper, properly credited and are the license and terms of use explicitly mentioned and properly respected?
    \item[] Answer: \answerYes{} 
    \item[] Justification: We discuss our data in \Cref{sec:method-data} and base model use in \Cref{sec:method}.

\item {\bf New assets}
    \item[] Question: Are new assets introduced in the paper well documented and is the documentation provided alongside the assets?
    \item[] Answer: \answerNA{} 
    \item[] Justification: 

\item {\bf Crowdsourcing and research with human subjects}
    \item[] Question: For crowdsourcing experiments and research with human subjects, does the paper include the full text of instructions given to participants and screenshots, if applicable, as well as details about compensation (if any)? 
    \item[] Answer: \answerNA{} 
    \item[] Justification: 

\item {\bf Institutional review board (IRB) approvals or equivalent for research with human subjects}
    \item[] Question: Does the paper describe potential risks incurred by study participants, whether such risks were disclosed to the subjects, and whether Institutional Review Board (IRB) approvals (or an equivalent approval/review based on the requirements of your country or institution) were obtained?
    \item[] Answer: \answerNA{} 
    \item[] Justification: 

\item {\bf Declaration of LLM usage}
    \item[] Question: Does the paper describe the usage of LLMs if it is an important, original, or non-standard component of the core methods in this research? Note that if the LLM is used only for writing, editing, or formatting purposes and does not impact the core methodology, scientific rigorousness, or originality of the research, declaration is not required.
    \item[] Answer: \answerYes{}
    \item[] Justification: We use gemini-3-flash-preview as the VLM reward
      model, which is a core component of our training pipeline
      providing the sole source of supervision. Its role is described
      in \Cref{sec:method-reward} and the full scoring prompts are
      provided in \Cref{app:reward-prompt}.


\end{enumerate}
\fi

\end{document}